\documentclass{egpubl}
\usepackage{eg2026}
 
\STAR                   
\CGFStandardLicense

\usepackage[T1]{fontenc}
\usepackage{dfadobe}  

\usepackage{cite}  
\BibtexOrBiblatex
\electronicVersion
\PrintedOrElectronic
\ifpdf \usepackage[pdftex]{graphicx} \pdfcompresslevel=9
\else \usepackage[dvips]{graphicx} \fi

\usepackage{egweblnk} 

\usepackage{color}
\usepackage{soul}
\usepackage{graphicx, animate}
\usepackage{tabularx}
\usepackage{wrapfig}

\usepackage{amssymb}
\usepackage{amsmath}
\usepackage[table,xcdraw]{xcolor}
\usepackage{tikz}
\usetikzlibrary{spy}
\usepackage[utf8]{inputenc}
\usepackage{pgfplots}
\pgfplotsset{compat=1.18}
\usepackage{pifont}
\usepackage{makecell}
\usepackage{float}
\usepackage{array}
\usepackage{longtable}
\usepackage{booktabs}
\usepackage{multicol}
\usepackage{multirow}
\usepackage[ruled]{algorithm2e} 

\SetAlFnt{\small}
\SetAlCapFnt{\small}
\SetAlCapNameFnt{\small}
\SetAlCapHSkip{0pt}
\usepackage{svg}

\newcommand{\etal}{\textit{et al.}}

\definecolor{title_purple}{rgb}{0.65,0.1,0.65}
\definecolor{darkred}{rgb}{0.7,0.1,0.1}
\definecolor{darkgreen}{rgb}{0.1,0.6,0.1}
\definecolor{green}{rgb}{0, 0.5, 0}
\definecolor{orange}{rgb}{0.8, 0.6, 0.2}
\definecolor{red}{rgb}{1.0, 0.0, 0.0}
\definecolor{teal}{rgb}{0.0, 0.4, 0.4}
\definecolor{purple}{rgb}{0.65,0.0,0.65}
\definecolor{saffron}{rgb}{0.95,0.75,0.2}
\definecolor{turquoise}{rgb}{0.0,0.5,0.5}
\definecolor{black}{rgb}{0.0, 0.0, 0.0}
\definecolor{gray}{rgb}{0.5, 0.5, 0.5}

\definecolor{gd_red}{rgb}{0.650980392,0.109803922,0}
\definecolor{gd_green}{rgb}{0.41568627450980394,00.6588235294117647,0.30980392156862746}
\definecolor{gd_blue}{rgb}{0.211765,0.423529,0.762353}

\newcommand{\revone}[1]{{\color{black}#1}}  

\title[Advances in Neural 3D Mesh Texturing: A Survey]%
      {Advances in Neural 3D Mesh Texturing: A Survey}

\author[S. R. K. Perla, H. Zhang, \& A. Mahdavi-Amiri]
{\parbox{\textwidth}{\centering 
{\hypersetup{colorlinks=true, urlcolor=black}\href{https://sairajk.github.io/}{Sai Raj Kishore Perla}\orcid{0009-0001-6819-8282}} $\quad$ 
{\hypersetup{colorlinks=true, urlcolor=black}\href{https://www.cs.sfu.ca/~haoz/}{Hao Zhang}\orcid{0000-0003-1991-119X}} $\quad$ 
{\hypersetup{colorlinks=true, urlcolor=black}\href{https://arash-mham.github.io/}{Ali Mahdavi-Amiri}\orcid{0000-0002-4693-3565}} 
        }
        \\
{\parbox{\textwidth}{\centering Simon Fraser University
       }
}
}

\begin{document}

\teaser{
 \includegraphics[width=0.95\linewidth]{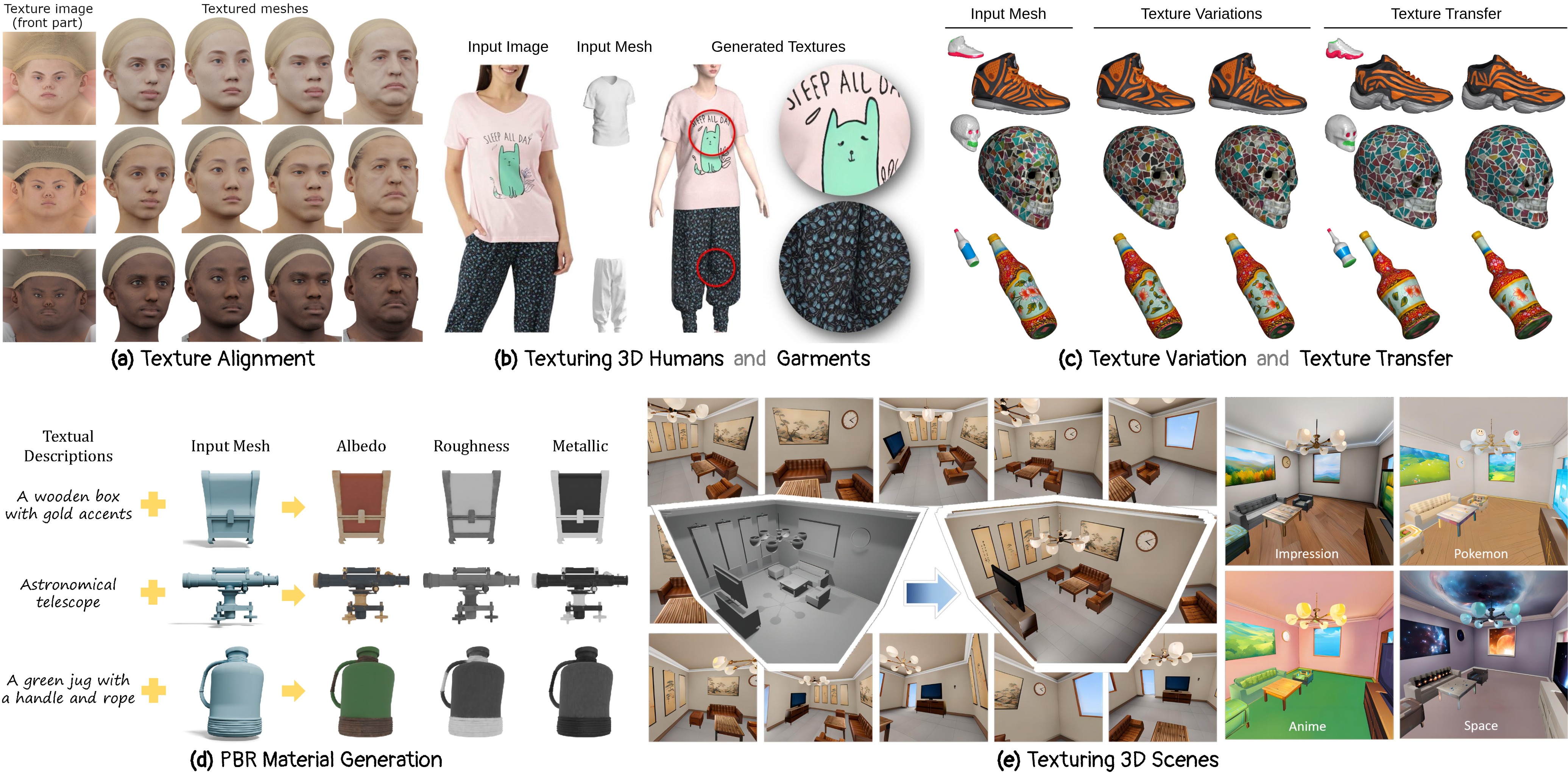}
 \centering
  \caption{
    \textbf{Neural 3D Mesh Texturing} spans diverse settings, with works addressing different aspects of surface appearance generation and control.
    For instance, (a) \emph{Texture Alignment}, learning class-consistent UV mappings~\cite{chen2022auvnet}; (b) \emph{Texturing Humans and Garments}~\cite{zhang2024fabricdiffusion}, also demonstrating texture synthesis from an input image; (c) \emph{Texture Variation and Transfer}~\cite{mitchel2024fieldlatents}, also illustrating texturing from an input textured mesh; (d) \emph{PBR Material Generation}~\cite{xiong2025texgaussian}, also showcasing texturing from a text prompt; and (e) \emph{Scene-level Texturing}~\cite{wang2024roomtex}.
    Figure adapted from~\cite{chen2022auvnet,zhang2024fabricdiffusion,mitchel2024fieldlatents,xiong2025texgaussian,wang2024roomtex}.
  }
\label{fig:teaser}
}

\maketitle

\begin{abstract}
%
Texturing 3D meshes plays a vital role in determining the visual realism of digital objects and scenes. Although recent generative 3D approaches based on Neural Radiance Fields and Gaussian Splatting can produce textured assets directly, polygonal meshes remain the core representation across modeling, animation, visual effects, and gaming pipelines. Neural 3D mesh texturing therefore continues to be an essential and active area of research. In this survey, we present a comprehensive review of recent advances in neural 3D mesh texturing, covering methods for texture synthesis, transfer, and completion. We first summarize key foundations in mesh geometry, texture mapping, differentiable rendering, and neural generative models, and then organize the literature into a unified taxonomy spanning early GAN-based methods to modern diffusion-based pipelines. We further analyze common architectures and supervision strategies, review datasets and evaluation protocols, and discuss emerging applications, practical/commercial systems, and open challenges. Together, these insights provide a structured perspective on the current landscape and help guide future developments in learning-based 3D mesh texturing.

\vspace{3pt}
\noindent
\emph{\textbf{Project Page:} {\href{https://sairajk.github.io/neural-mesh-texturing/}{\texttt{sairajk.github.io/neural-mesh-texturing}}}}
\vspace{2pt}


\end{abstract}  


\section{Introduction}
\label{sec:intro}

Texturing 3D meshes has long been a central task in computer graphics and digital content creation, providing the visual richness that transforms geometric models into realistic and stylistically expressive assets~\cite{heckbert1986mapping,dischler2001survey3dtexturing}. In traditional pipelines, this process typically relies on manual UV unwrapping together with artist-driven material setup and texture authoring~\cite{sheffer2006survey,levy2002lscm,sheffer2005abfpp}. These steps usually require artistic skill and extensive labor. While procedural and image/example-based texturing techniques have eased some of this burden~\cite{pietroni2010solidtexture,weinhaus1997texturemapping,turk2001surfacestexture,wei2009texturestar}, they remain largely constrained by handcrafted priors and limited semantic understanding, making it difficult to synthesize complex object appearances across diverse categories.

Recent advances in neural visual processing have shown great promise and significantly reshaped this landscape (Fig.~\ref{fig:teaser}). By leveraging deep learning, differentiable rendering~\cite{ravi2020pytorch3d,jatavallabhula2019kaolin}, and large-scale image generative models~\cite{rombach2022ldm}, recent systems can automatically generate, complete, or refine textures for 3D meshes~\cite{chen2023text2tex,perla2024easitex}. Instead of relying solely on manual texture authoring, these methods combine geometric reasoning with learned visual priors to guide texture synthesis from text, images, or sparse observations. This has opened a new paradigm for 3D asset creation, enabling meshes to be textured with greater realism, consistency, and flexibility while substantially reducing manual effort.

An early line of neural texturing methods relied on adversarial and feed-forward generation, learning to synthesize textures from weak 2D supervision through differentiable rendering~\cite{yu2021learning_texture_generators,siddiqui2022texturify,bokhovkin2023mesh2tex}. These works demonstrated that neural networks could directly generate plausible mesh textures, but were often limited in semantic control, generalization, and diversity. Subsequent approaches shifted toward optimization-based pipelines guided by pretrained vision-language models such as CLIP~\cite{michel2022text2mesh,ma2023xmesh,chen2022tango,radford2021clip}, showing that large-scale 2D priors can effectively steer texture synthesis in 3D. Later, diffusion-based optimization methods~\cite{chen2023fantasia3d,yeh2024texturedreamer,poole2023dreamfusion} further improved fidelity and diversity through Score Distillation Sampling (SDS)~\cite{poole2023dreamfusion} and related guidance schemes~\cite{wang2023prolificdreamer}. More recently, iterative and feed-forward diffusion pipelines~\cite{chen2023text2tex,richardson2023texture,tang2024makeitvivid} have enabled faster synthesis of high-resolution textures with stronger multi-view coherence. Complementary advances have also targeted human avatar texturing~\cite{nam2025parte,liu2024texdreamer}, physically based material generation~\cite{huang2025material,chen2022tango}, and local or procedural control~\cite{decatur2024paintbrush}, further expanding the automation and editability of neural mesh texturing.

Despite these breakthroughs, challenges remain. Neural texturing still struggles with efficiency, multi-view consistency, and physical realism under varying illumination. Moreover, the scarcity of high-quality textured datasets and standardized evaluation mechanisms complicates benchmarking and comparison across methods. As the field moves toward prompt-driven 3D content generation and broader adoption of neural methods for content creation and manipulation, understanding the assumptions, algorithmic setups, the network architectures underlying these methods, as well as the trade-offs among them, becomes increasingly important.

To the best of our knowledge, while several surveys discuss texture synthesis and mapping, none focus exclusively on \emph{neural 3D mesh texturing} as we do. In 2D, surveys of exemplar- and patch-based texture synthesis focus on image domains rather than 3D surfaces~\cite{barnes2017patchsynthsurvey,raad2018exemplarsurvey,bhunia2019TexRet}. In 3D, prior surveys have reviewed non-neural texture mapping and texturing topics including parameterization and mapping~\cite{heckbert1986mapping,sheffer2006survey}, texture mapping from photographs~\cite{weinhaus1997texturemapping}, example-based texture synthesis on surfaces~\cite{turk2001surfacestexture}, and 3D texturing more broadly~\cite{dischler2001survey3dtexturing}. Later surveys consolidated example-based and volumetric texture synthesis, but they do not cover deep-learning-based techniques~\cite{wei2009texturestar,pietroni2010solidtexture}. Relevant recent surveys on neural generation typically address broader themes, such as text-guided 3D editing~\cite{lu2024textguided3deditingsurvey}, neural stylization~\cite{chen2025neuralstylizationsurvey}, or text-to-3D generation~\cite{li2023text3dsurvey}.

In this survey, we review recent advances in \emph{neural 3D mesh texturing}, focusing on neural methods that operate directly on 3D meshes. For the purposes of this survey, we view 3D mesh texturing as involving two tightly coupled subproblems: \emph{(i) surface parameterization}, which defines a 2D signal domain on the mesh surface (\textit{e.g.}, UV charts and seams) where appearance can be stored and sampled, and \emph{(ii) texture synthesis}, which generates the actual appearance representation, such as RGB textures, material maps, or learned features, either in that domain or in view space before baking the result back onto the mesh. Given the breadth of mesh parameterization research, this survey focuses primarily on \emph{neural texture synthesis} and discusses parameterization only where it directly affects texture representation and learning; we refer readers to dedicated surveys for a more comprehensive overview~\cite{sheffer2006survey,hormann2008meshparam,floater2005spa}.

We organize the literature primarily by recurring methodological families, while also reflecting the field’s evolution from early foundational neural methods—including GAN-based and weakly supervised pipelines—to optimization-based approaches, such as vision--language-guided and diffusion-guided optimization, and, more recently, accelerated diffusion-based methods that synthesize textures through iterative, synchronized multi-view, or feed-forward generation. We also discuss specialized domains such as 3D human texturing and commercial systems, along with datasets and evaluation metrics, highlighting gaps and open challenges in the current research landscape. Ultimately, this survey aims to provide a comprehensive perspective on how neural models are transforming 3D texture generation by combining artistic creativity, physical realism, and scalable automation.

In the following, we first introduce key preliminaries in Sec.~\ref{sec:preliminaries}, followed by a discussion of guidance signals used in neural mesh texturing in Sec.~\ref{sec:guidance}. Sec.~\ref{sec:neural-texturing} surveys texturing methods, including foundational neural pipelines, optimization-based methods, and accelerated diffusion-based approaches, as well as 3D human texturing and commercial systems. We then review datasets and evaluation metrics in Sec.~\ref{sec:datasets-metrics}, summarize applications in Sec.~\ref{sec:applications}, discuss limitations in Sec.~\ref{sec:limitations}, and conclude with future work in Sec.~\ref{sec:future-work}.

\section{Neural Texturing Overview and Background}
\label{sec:preliminaries}


This section outlines the background required to understand and compare methods for neural 3D mesh texturing. We review the fundamentals of 3D mesh representation in Sec.~\ref{subsec:prelim-meshes}, introduce surface parameterization as a bridge from 3D surfaces to 2D texture domains in Sec.~\ref{subsec:prelim-param}, summarize rendering and differentiable rendering in Sec.~\ref{subsec:prelim-diffrender}, discuss physically based rendering (PBR) and PBR materials in Sec.~\ref{subsec:pbr}, and outline key learning paradigms relevant to neural mesh texturing—neural fields~\cite{xie2022neuralfields}, vision--language models~\cite{radford2021clip,li2022blip}, VAEs~\cite{kingma2019introvae}, GANs~\cite{goodfellow2014gan}, and diffusion models~\cite{ho2020ddpm}—in Sec.~\ref{subsec:prelim-neural}. Together, these topics provide the foundation for the survey that follows.

\subsection{Why 3D Meshes?}
\label{subsec:prelim-meshes}

\begin{figure}[t]
  \centering
  \includegraphics[width=0.99\linewidth]{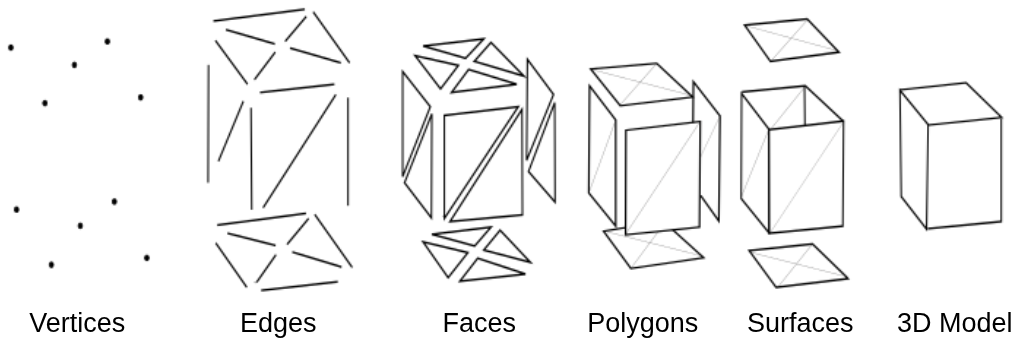}
  \caption{A 3D mesh is defined by vertices (points), edges (line segments), and faces (surface elements, often triangles or quads), which combine into surface patches that approximate the object’s geometry. Figure reproduced from~\cite{mesh_overview_commons_2009}.}
  \label{fig:mesh_basics}
\end{figure}

\begin{figure}[t]
  \centering
  \includegraphics[width=0.99\linewidth]{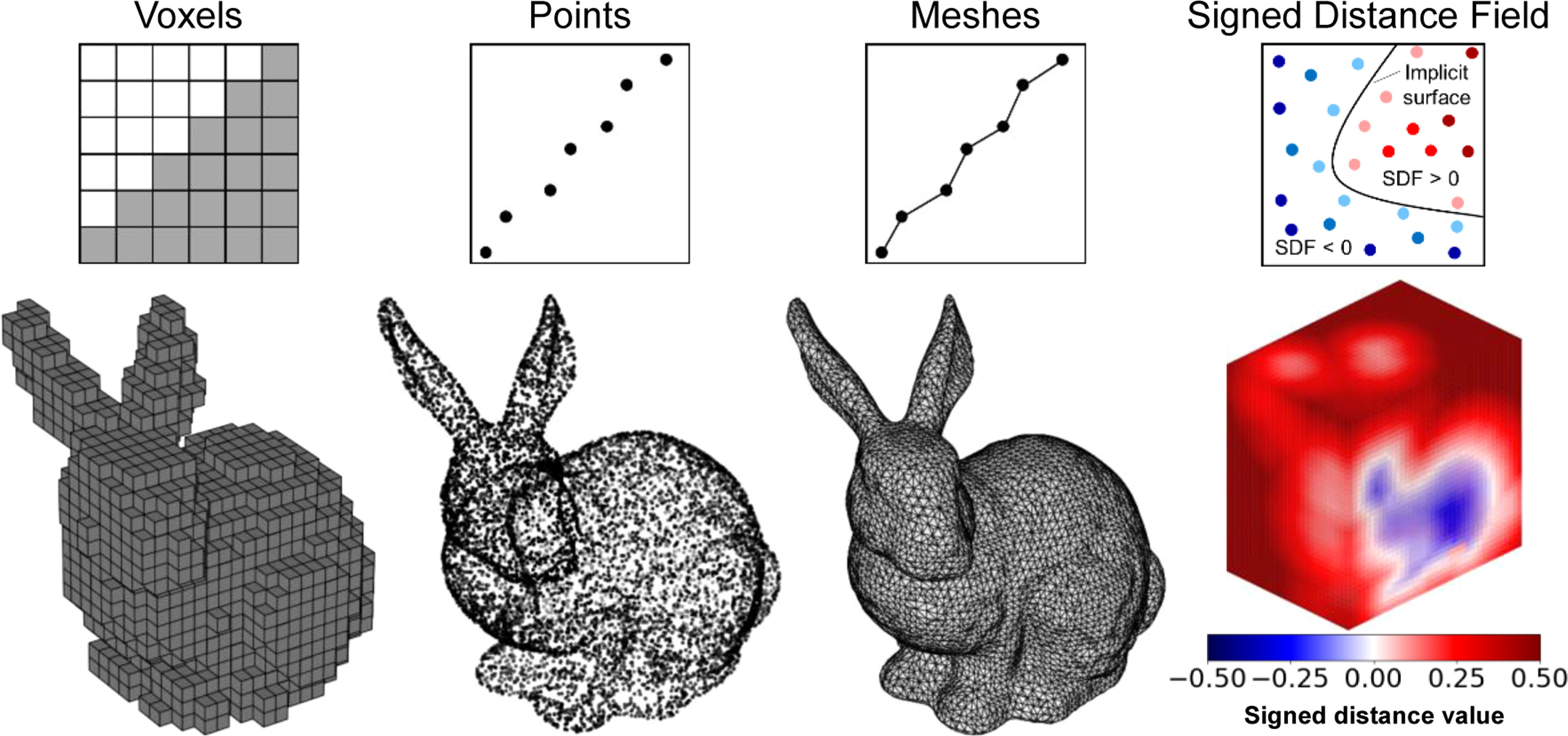}
  \caption{3D shape representations. Voxels discretize space, point clouds sample points, meshes encode surface connectivity, and SDFs represent shapes implicitly via a continuous signed distance function. Figure reproduced from~\cite{attar2023_3dcomparison}.}
  \label{fig:3d_rep_comparison}
\end{figure}

Triangle meshes are the dominant explicit representation for surfaces in computer graphics and vision~\cite{botsch2010pmp,akenine2018rtr4,deitke2023objaverse,koch2019abc}. A mesh encodes geometry as a finite set of vertex positions together with a connectivity structure of edges and faces, which defines adjacency relations and surface topology (Fig.~\ref{fig:mesh_basics}). In practice, triangle meshes support rich per-primitive attributes such as normals, colors, and material identifiers, making them a natural carrier for appearance information~\cite{akenine2018rtr4}. From a texturing standpoint, this explicit representation is crucial: appearance channels can be deterministically associated with well-defined surface elements, and mesh connectivity supports the consistent propagation of these signals across the surface~\cite{botsch2010pmp}.

Contrasting meshes with alternative shape encodings (Fig.~\ref{fig:3d_rep_comparison}) helps explain why they are the primary surface representation considered in this survey~\cite{botsch2010pmp}. Point-based and surfel models capture geometry without explicit connectivity and are attractive for point- or splat-based rendering, but they do not provide a standard 2D surface parameterization for texture mapping~\cite{pfister2000surfels,rusinkiewicz2000qsplat}. Volumetric grids---dense or sparse---have been widely used for image-to-shape prediction and 3D reconstruction, but become memory-bound at high resolutions and typically require isosurface extraction to yield deployable surfaces~\cite{choy2016r2n2,lorensen1987marchingcubes}. Continuous implicit fields represent shapes as decision boundaries or signed-distance zero sets~\cite{mescheder2019occupancy,park2019deepsdf}, while radiance fields directly model view-dependent appearance in 3D space~\cite{mildenhall2020nerf}. Although these formulations excel at reconstruction and novel-view synthesis, their outputs are often converted to triangle meshes (\textit{e.g.}, via marching cubes~\cite{lorensen1987marchingcubes}) to interoperate with texture-map-based tools and rendering engines.

\begin{figure}[t]
  \centering
  \includegraphics[width=0.99\linewidth]{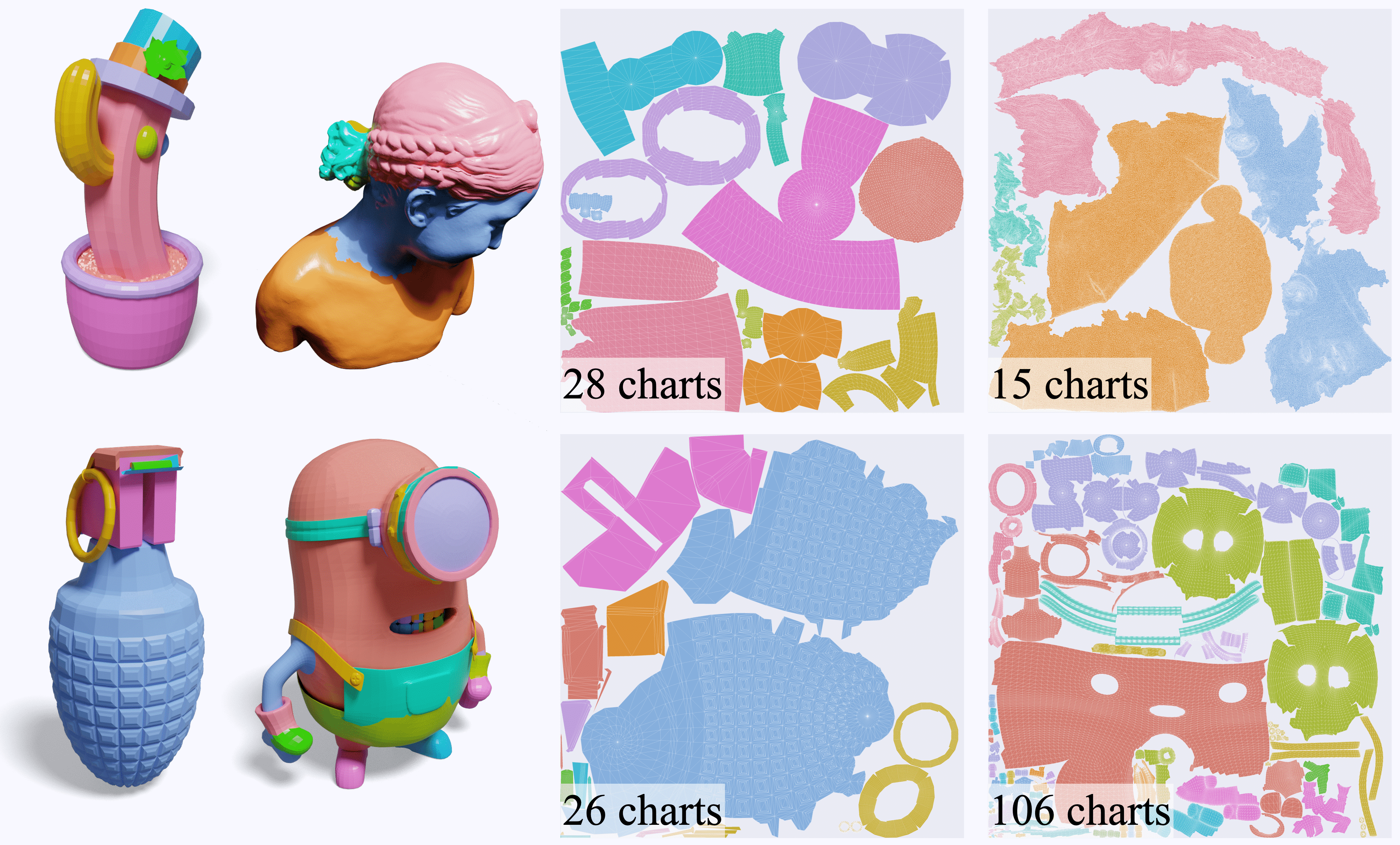}
  \caption{
  Mesh parameterization maps a 3D surface to a 2D domain (UV space) by partitioning it into charts and flattening each into a planar region. Example meshes (left) and UV layouts (right) show varying chart counts. Figure adapted from~\cite{wang2025partuv}.
  }
  \label{fig:mesh_param}
\end{figure}

\subsection{Mesh Parameterization}
\label{subsec:prelim-param}

Mesh parameterization assigns each point on a surface a coordinate in a two-dimensional domain, defining a mapping from 3D surface space to 2D~\cite{sheffer2006survey} (Fig.~\ref{fig:mesh_param}). In practice, because general surfaces typically cannot be flattened into a single planar domain without introducing cuts and/or distortion, a triangle mesh is partitioned into a set of charts—connected surface patches—each mapped to a planar region; the collection of these charts, packed into one or more 2D texture images, forms an atlas. The resulting per-vertex UV coordinates define the texture parameterization and are interpolated per fragment during rendering. Charts meet along seams where the UV mapping is discontinuous. Parameterization also enables raster textures to be sampled and filtered during rendering. To improve filtering across scales, textures are typically stored in a mipmap hierarchy, which precomputes progressively downsampled versions of the image to reduce aliasing during minification~\cite{williams1983pyramidal}.

From a texturing standpoint, parameterization decouples geometry from appearance. Appearance attributes---such as albedo, normal, and roughness---can be stored in image space at resolutions largely independent of mesh tessellation and sampled consistently during rendering~\cite{yuksel2019rethink}. Once parameterized, texture resolution can be increased (\textit{e.g.}, by using higher-resolution UV images) without altering the underlying geometry, allowing appearance fidelity to scale with available storage and bandwidth budgets.

While traditional pipelines represent appearance as one or more raster texture images in UV space (\textit{e.g.}, albedo/normal/roughness maps), some neural methods replace or augment these images with \emph{learned} representations. A representative example is \emph{neural textures}---high-dimensional feature maps stored on a mesh (often still in UV space) and optimized jointly with a neural renderer, enabling the renderer to decode view-dependent appearance from learned features rather than directly from RGB texels~\cite{thies2019deferred}. Other works represent texture implicitly as a continuous function over surface coordinates or 3D points (Sec.~\ref{sssec:neural_fields}), but parameterization (UVs, seams, and chart layout) remains central whenever supervision or outputs are baked back into a 2D atlas.

Several alternatives to UV-based parameterization exist. \textit{Per-face texturing} (\textit{e.g.}, Ptex~\cite{burley2008ptex}) stores signals on individual faces without requiring a global UV layout, reducing seam management and atlas packing overhead, but it can introduce discontinuities when filtering across face boundaries. \textit{Per-vertex colors} are simple and compact, but the achievable detail is constrained by mesh resolution (vertex density) and connectivity (triangle distribution), and interpolation across faces can blur high-frequency signals~\cite{akenine2018rtr4}. \textit{Volumetric or procedural textures} avoid UV maps entirely, but they introduce challenges in authoring, editing, and manual control. Consequently, in practice, UV-based parameterization remains the most portable representation across offline and real-time rendering pipelines and continues to serve as the standard substrate for current texture libraries and authoring workflows~\cite{yuksel2019rethink}.

Parameterization itself is also an active research problem that can materially affect downstream texturing quality, since distortion, seam placement, and chart fragmentation can influence both downstream learning behavior and the visibility of seams after synthesis. Classical methods target conformality and distortion control (\textit{e.g.}, least-squares conformal maps~\cite{levy2002lscm}, angle-based flattening~\cite{sheffer2005abfpp}, and boundary-first parameterization~\cite{sawhney2017bff}), often coupled with seam optimization to reduce fold-overs and atlas fragmentation~\cite{li2018optcuts}. More recently, learned and task-adaptive approaches have emerged that automate or adapt parameterization toward data-driven objectives~\cite{low2022mna,liu2023dawand,wang2025partuv}: \emph{Flatten Anything} proposes an unsupervised neural architecture that learns free-boundary surface parameterizations and can adaptively infer reasonable cuts and UV boundaries even from unstructured surface samples~\cite{zhang2024flatten}, while \emph{Auto-Regressive Surface Cutting} (SeamGPT) formulates seam generation as next-token prediction to produce semantically cleaner, less fragmented seam layouts for UV unwrapping~\cite{li2025seamgpt}. Such methods complement robust classical solvers by reducing manual charting effort and by producing atlases better matched to modern texturing pipelines.

For these reasons, this survey considers mesh parameterization a foundational step for neural 3D mesh texturing: it decouples geometry from appearance by defining a 2D signal domain in which textures are stored (as images or learned features), optimized or regularized during learning, and consumed by renderers~\cite{yuksel2019rethink,sander2001tmpm,williams1983pyramidal}. Given the breadth of mesh parameterization research, we limit our discussion to aspects most relevant to texture representation and learning, and refer readers to dedicated surveys for a comprehensive overview~\cite{sheffer2006survey,hormann2008meshparam,floater2005spa}.

\subsection{Differentiable Mesh Rendering}
\label{subsec:prelim-diffrender}
Rendering converts a 3D scene into a 2D image by simulating what a camera would observe. Given a triangle mesh with associated materials and lighting, a forward rendering pipeline determines which surface points are visible and computes their appearance in the image. In rasterization-based rendering, mesh vertices are transformed from object space to camera space, projected onto the image plane, assembled into screen-space primitives, and resolved for visibility using a depth buffer~\cite{akenine2018rtr4}. In ray- or path-traced rendering, rays are cast through the scene to determine visible surface points and their light transport interactions~\cite{pharr2023pbrt4}. The visible points are then shaded by evaluating a material model under incident illumination~\cite{kajiya1986re}, while textures are sampled---with filtering to reduce aliasing---to provide spatially varying parameters such as color or roughness~\cite{pharr2023pbrt4}. The resulting pixel values are finally composited with a background (solid color, image, or environment map) and written to an output image or framebuffer.

\begin{figure}[t]
  \centering
  \includegraphics[width=0.99\linewidth]{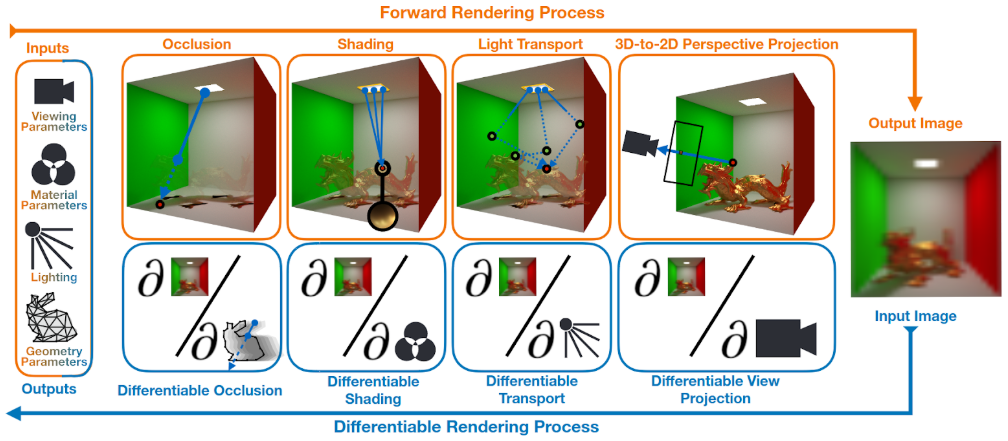}
  \caption{
  Differentiable rendering enables backpropagation of image-space losses through the rendering pipeline (occlusion, shading, light transport, and projection) to optimize scene parameters (geometry, materials, lighting, and viewpoint). Figure reproduced from~\cite{bannister2025dvc}.
  }
  \label{fig:diff_render}
\end{figure}

Differentiable rendering makes parts of the rendering process amenable to gradient-based optimization, enabling end-to-end learning of geometry, materials, and textures from image supervision (Fig.~\ref{fig:diff_render}). A central challenge is the discontinuities inherent to visibility and rasterization. Classical approaches address this by introducing differentiable approximations or surrogate gradients, as in OpenDR~\cite{loper2014opendr}, or by reformulating rasterization to expose stable derivatives, as in Neural Mesh Renderer~\cite{kato2018nmr}, Soft Rasterizer~\cite{liu2019softras}, and analytical treatments such as DIB-R~\cite{chen2019dibr}. In path-traced settings, differentiability is achieved via Monte Carlo estimators that handle visibility boundaries and other discontinuities~\cite{li2018edgesampling,loubet2019reparam,zhang2020pathspace,pharr2023pbrt4}. These advances are embodied in modern research renderers and toolchains: Mitsuba 2/3~\cite{nimier2019mitsuba2} with its Dr.Jit compiler~\cite{jakob2022drjit} supports forward- and reverse-mode differentiation for full light transport, while rasterization-based back ends such as nvdiffrast~\cite{laine2020nvdiffrast} and deep-learning libraries like PyTorch3D~\cite{ravi2020pytorch3d} and Kaolin~\cite{jatavallabhula2019kaolin} provide efficient GPU pipelines for differentiable rendering.


\subsection{Physically Based Rendering and PBR Material Maps}
\label{subsec:pbr}

Physically based rendering (PBR) aims to model light--material interaction using physically grounded scattering and transport principles, so that appearance remains predictable under changing illumination and viewpoints. In the rendering equation formulation~\cite{kajiya1986re}, outgoing radiance is determined by emitted radiance together with incident radiance modulated by the surface scattering function. For opaque surfaces, this scattering is typically modeled by a \emph{bidirectional reflectance distribution function} (BRDF), and more generally by a BSDF when transmission is included~\cite{pharr2023pbrt4,veach1997thesis}. The BRDF describes how incident light from one direction is reflected into another at a surface point~\cite{nicodemus1977reflectance}. In modern PBR systems, this reflectance is commonly modeled using energy-conserving microfacet models~\cite{cook1981reflectance,burley2012disney,pharr2023pbrt4}. This perspective distinguishes \emph{geometry} (shape and surface normals) from \emph{material} parameters that govern diffuse and specular response, Fresnel effects~\cite{schlick1994brdf,born1999principles}, and the distribution of micro-surface orientations.

\begin{figure}[t]
  \centering
  \includegraphics[width=0.99\linewidth]{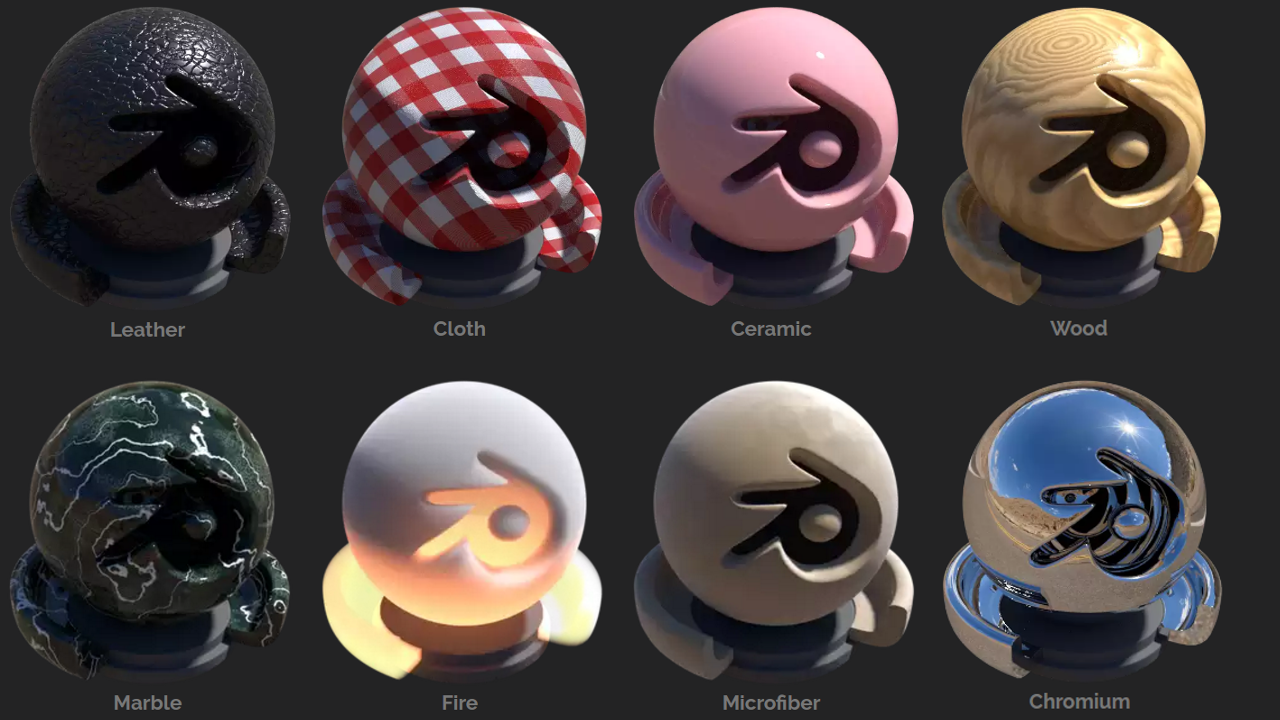}
  \caption{
  Representative PBR materials rendered under identical geometry and lighting. Varying material map parameters (\textit{e.g.}, base color/albedo, roughness, metallic, normals, and emissive terms) yield distinct, relightable appearances such as leather, cloth, ceramic, wood, marble, microfiber, and chromium, illustrating how PBR materials extend beyond a single RGB texture. Figure reproduced from~\cite{wolf2016pbrmaterialsaddon}.
  }
  \label{fig:pbr_materials}
\end{figure}

A single RGB texture is often treated as a ``color'' signal (\textit{e.g.}, diffuse/albedo) and may implicitly bake lighting, specular highlights, or other view-dependent effects, limiting relightability. In contrast, a PBR material is typically represented by a \emph{set of texture maps} in UV space that parameterize a shading model. Modern real-time pipelines commonly adopt a metallic--roughness parameterization (as in Khronos glTF), where \texttt{baseColor} serves as diffuse albedo for dielectrics (non-metals) and specular reflectance color for conductors (metals), while \texttt{metallic} and \texttt{roughness} control reflectance behavior and highlight sharpness~\cite{khronosPBR,filamentMaterials,gltf20spec}. Additional maps such as normal, ambient occlusion, emissive, and, in broader PBR workflows, height/displacement can further enrich appearance; optional extensions such as clearcoat, sheen, transmission, and volume provide additional control over layered and transmissive effects~\cite{filamentMaterials,khronosPBR} (Fig.~\ref{fig:pbr_materials}). We use \emph{PBR material generation} to refer to methods that explicitly synthesize such multi-map material representations rather than only a single RGB texture.

Generating high-quality PBR materials is harder than predicting a single RGB texture because multiple channels must be both \emph{individually plausible} and \emph{mutually consistent}. First, material maps are tightly coupled: normal detail should be consistent with roughness and specular response, and metallic regions should exhibit physically consistent baseColor/reflectance behavior~\cite{burley2012disney,filamentMaterials}. Second, disentangling intrinsic material from illumination is ill-posed when supervision is limited to rendered RGB images, often causing ``baked-in'' shading artifacts that break under relighting. Third, real materials can be spatially varying and layered (paint/clearcoat, fabrics, skin/hair), and their appearance depends on the target shader and renderer conventions (color space, parameter ranges, map packing), making cross-system generalization nontrivial~\cite{filamentMaterials,khronosPBR}. Finally, large-scale paired datasets of geometry with calibrated, multi-map ground-truth materials remain comparatively scarce relative to RGB imagery, complicating evaluation and often encouraging simplified supervision or incomplete material factorization~\cite{deschaintre2018opensvbrdf}. These issues motivate current research directions that incorporate geometry cues, multi-view consistency, priors from image generative models, and renderer-aware training to better support relightable, editable PBR materials.

\subsection{Neural Network Paradigms}
\label{subsec:prelim-neural}

Many neural mesh texturing pipelines can be understood through a small set of recurring paradigms for representing and generating surface appearance. We briefly introduce these paradigms---predicting textures or material maps in UV space, modeling appearance as a continuous function over the surface, and generating view-space images or latents that are later baked onto the mesh---as background for the methods surveyed in later sections.

\subsubsection{Neural Fields}
\label{sssec:neural_fields}

\begin{figure}[t]
  \centering
  \includegraphics[width=0.99\linewidth]{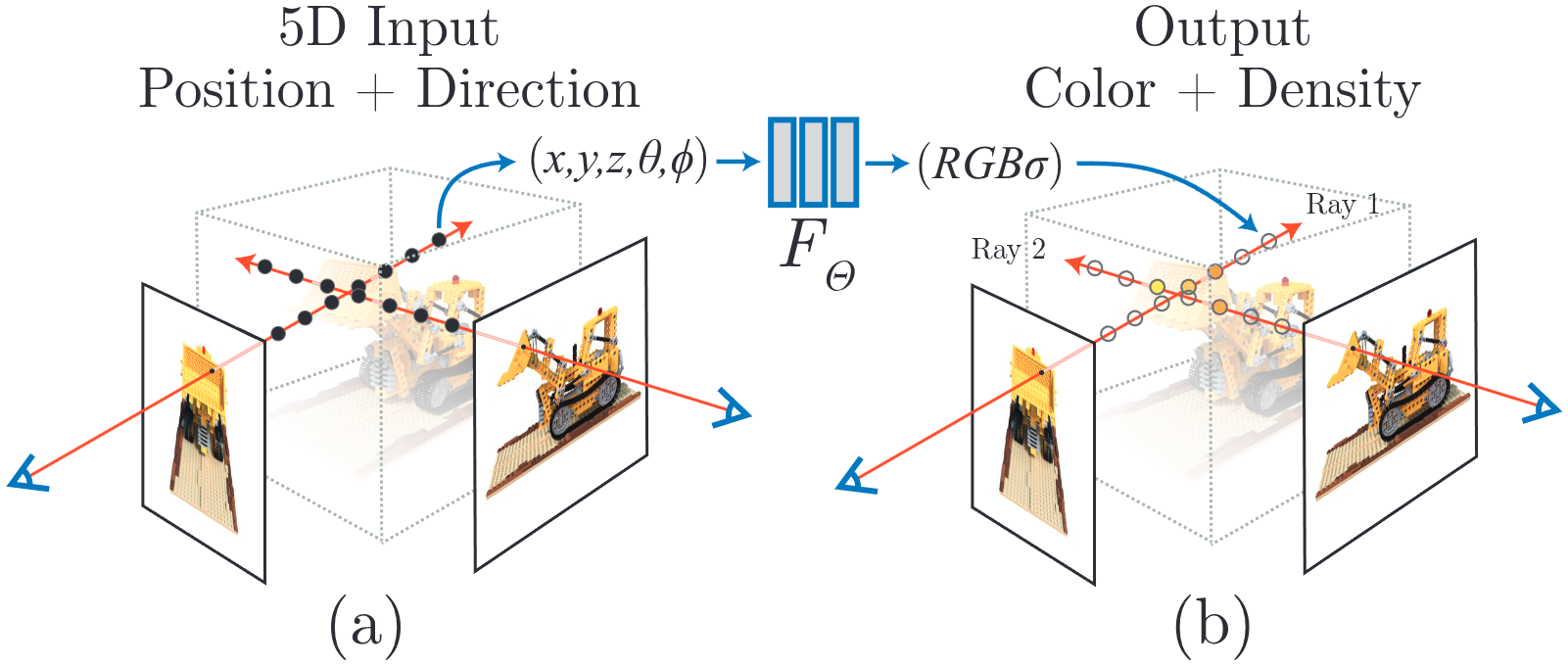}
  \caption{
  An overview of Neural Radiance Fields (NeRFs). Images are synthesized by sampling 5D coordinates—3D position $(x,y,z)$ and viewing direction $(\theta,\phi)$—along camera rays (a) and feeding them to a neural network that outputs color $(R,G,B)$ and volume density $\sigma$ (b).
  Figure adapted from~\cite{mildenhall2020nerf}.
  }
  \label{fig:nerf}
\end{figure}

Neural fields, also referred to as coordinate-based neural representations, model signals as continuous functions of space, optionally extended with direction, time, or other conditioning variables, and parameterized by a neural network. Formally, a neural field is a function
\begin{equation*}
    f_{\theta} : \mathbb{R}^{n} \to \mathbb{R}^{m}
\end{equation*}
that maps a coordinate---such as a 3D point \(\mathbf{x} \in \mathbb{R}^3\) or a tuple \((\mathbf{x}, \mathbf{d})\) with \(\mathbf{d} \in \mathbb{S}^2\) (\textit{i.e.}, a unit vector in \(\mathbb{R}^3\)) denoting a viewing direction---to a value representing signed distance, occupancy, color, or material parameters~\cite{xie2022neuralfields}. Unlike classical, explicitly sampled representations (\textit{e.g.}, images, voxel grids, or vertex-attached attributes on meshes), a neural field encodes content in network weights and can be queried at arbitrary spatial resolution. This formulation yields a compact, differentiable, and resolution-independent representation well suited to inverse problems from images~\cite{mescheder2019occupancy,park2019deepsdf}. Neural fields can be optimized end-to-end from image supervision via differentiable rendering. Consequently, Neural Radiance Fields (NeRFs) instantiate this idea for view synthesis by learning a mapping
\begin{equation*}
    f_{\theta}(\mathbf{x}, \mathbf{d}) \mapsto (\sigma, \mathbf{c}),
\end{equation*}
where \(\sigma\) denotes volume density and \(\mathbf{c}\) the view-dependent color (Fig.~\ref{fig:nerf}). Images are rendered via volumetric integration along camera rays~\cite{mildenhall2020nerf}.

Despite their strengths, neural fields present several challenges for texturing. Training and optimization can remain computationally expensive, and the learned signal may entangle lighting and material properties unless these factors are explicitly modeled, causing baked textures to capture specular highlights or shadows as albedo. In practice, neural fields and mesh-based textures are often complementary: fields excel at recovering and regularizing fine-scale appearance from images, while meshes and UV maps provide an editable and interoperable substrate for downstream use.


\subsubsection{Vision Language Models}
\label{subsec:vlms}

\begin{figure}[t]
  \centering
  \includegraphics[width=0.99\linewidth]{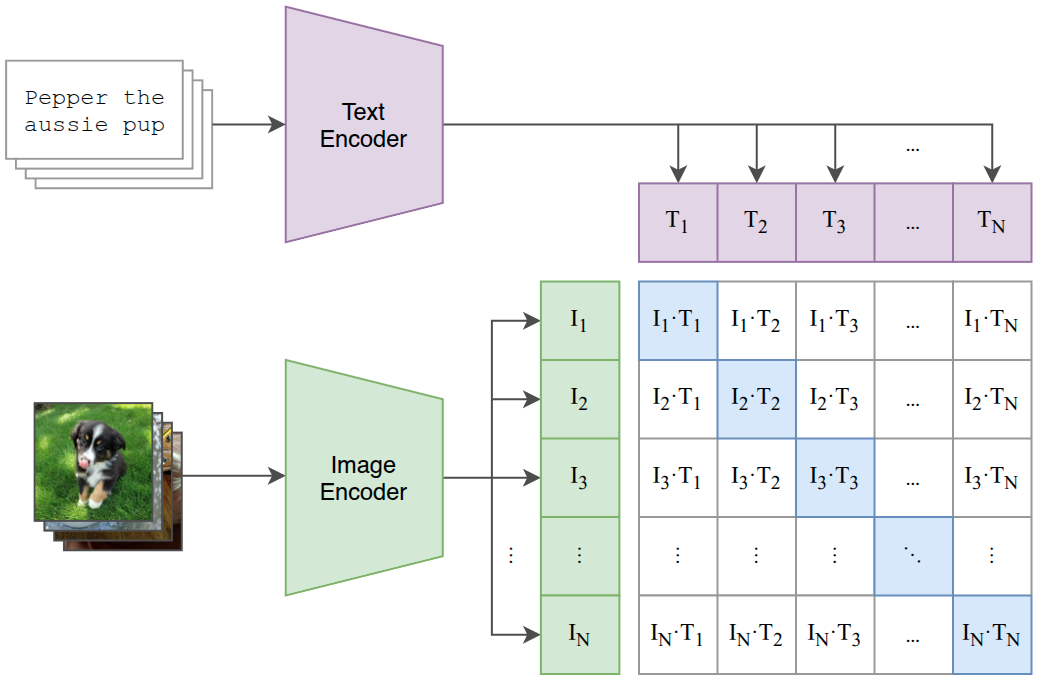}
  \caption{
  CLIP jointly trains image and text encoders using a contrastive loss to match paired images and texts within a batch. Figure adapted from~\cite{radford2021clip}.
  }
  \label{fig:clip_train}
\end{figure}

Vision--language models (VLMs) are neural networks trained on paired image--text data to align visual content with natural language. In contrast to vision-only encoders (which map images to features) or language-only models (which operate purely over text), VLMs learn shared cross-modal representations that support semantic alignment between modalities. In neural 3D mesh texturing, VLMs are especially valuable because they provide (i) open-vocabulary guidance for text-driven appearance synthesis, (ii) multi-view scoring of renders that links textual intent to visual fidelity, and (iii) region- or phrase-level grounding to target specific mesh parts.

A foundational instance is \emph{CLIP}~\cite{radford2021clip}, which learns dual encoders for images and text using a symmetric contrastive objective over large-scale web image--text pairs (Fig.~\ref{fig:clip_train}). By maximizing cosine similarity for matched image--text embeddings and minimizing it for mismatched pairs, CLIP induces a shared embedding space that supports robust zero-shot recognition and has become a widely used source of supervision for text-driven stylization and texture transfer~\cite{michel2022text2mesh,chen2022tango,khalid2022clipmesh}. \emph{GLIP}~\cite{li2022glip} extends this cross-modal alignment toward object detection and phrase grounding, producing region-aware, language-conditioned visual representations. \emph{BLIP}~\cite{li2022blip} further introduces a multimodal encoder--decoder framework that supports both understanding-oriented objectives (e.g., image--text matching) and generation tasks such as captioning and VQA.


\subsubsection{Variational Autoencoders}
\label{sssec:vae}

\begin{figure}[t]
  \centering
  \includegraphics[width=0.99\linewidth]{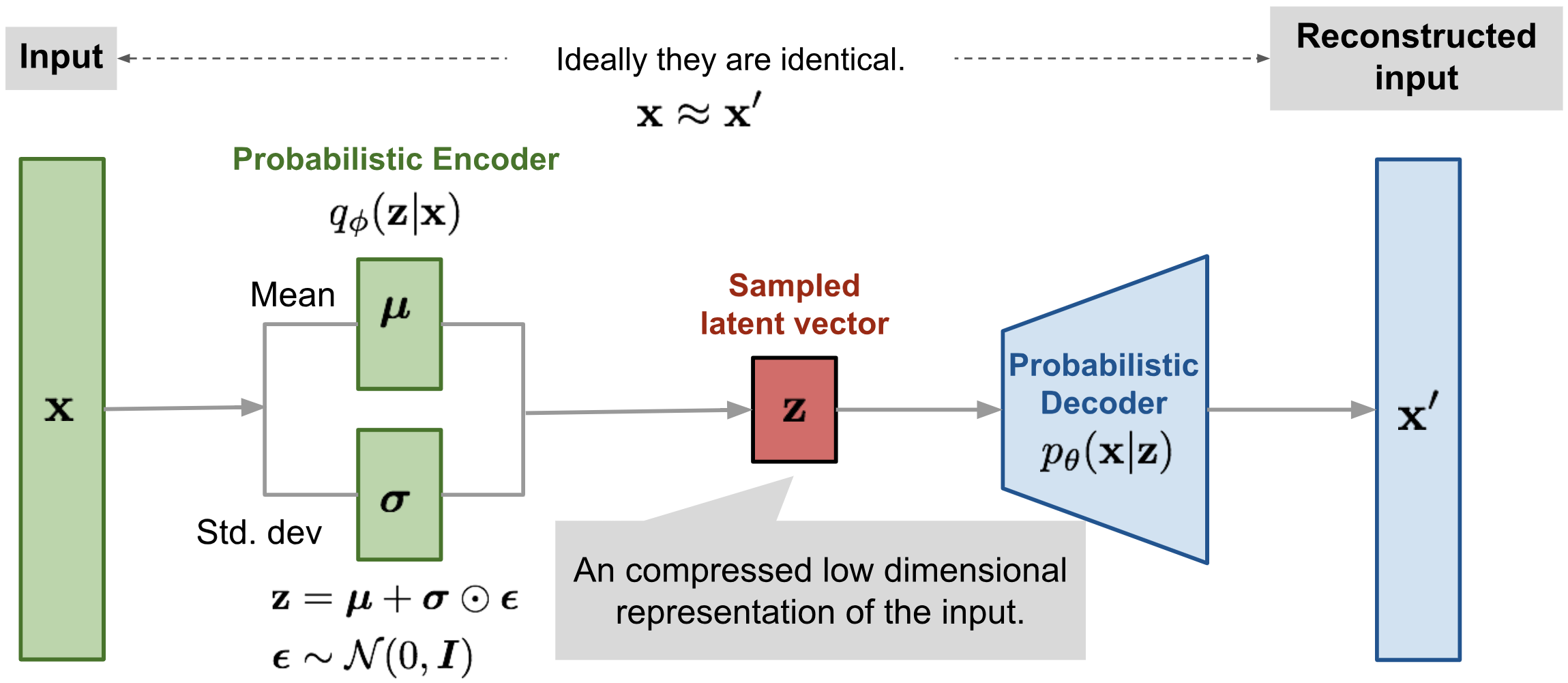}
  \caption{
  %
  %
  An overview of variational autoencoders (VAEs). A probabilistic encoder $q_\phi(\mathbf{z}\mid\mathbf{x})$ maps an input $\mathbf{x}$ to a Gaussian latent distribution parameterized by $(\mu,\sigma)$; a latent sample $\mathbf{z}=\mu+\sigma\odot\epsilon$ with $\epsilon\sim\mathcal{N}(\mathbf{0},\mathbf{I})$ is then decoded by $p_\theta(\mathbf{x}\mid\mathbf{z})$ to reconstruct $\mathbf{x}'\approx\mathbf{x}$. The model is trained by maximizing the evidence lower bound (ELBO) $\mathcal{L}_{\text{ELBO}}$ (see Sec.~\ref{sssec:vae}). Figure reproduced from~\cite{weng2018VAE}.
  }
  \label{fig:vae}
\end{figure}

Variational autoencoders (VAEs)~\cite{kingma2014vae} are latent-variable generative models that pair an encoder $q_{\phi}(\mathbf{z}\mid\mathbf{x})$ with a decoder $p_{\theta}(\mathbf{x}\mid\mathbf{z})$ and are trained by maximizing the evidence lower bound (ELBO)~\cite{jordan1999variational,neal1998emview,blei2017variational} \revone{(Fig.~\ref{fig:vae})}. For a single observation $\mathbf{x}\in\mathcal{X}$ and a latent code $\mathbf{z}\in\mathbb{R}^{d}$ with prior $p(\mathbf{z})$, the ELBO is
\[
\mathcal{L}_{\text{ELBO}}(\theta,\phi;\mathbf{x})
=\mathbb{E}_{q_{\phi}(\mathbf{z}\mid \mathbf{x})}\!\big[\log p_{\theta}(\mathbf{x}\mid \mathbf{z})\big]
-\mathrm{KL}\!\left(q_{\phi}(\mathbf{z}\mid \mathbf{x})\,\|\,p(\mathbf{z})\right),
\]
where $\mathrm{KL}(\cdot)$ denotes the Kullback--Leibler divergence~\cite{kullback1951information}. After training, generation proceeds by ancestral sampling: one first draws $\mathbf{z}\sim p(\mathbf{z})$ (\textit{e.g.}, $\mathcal{N}(\mathbf{0},\mathbf{I})$) and then samples $\mathbf{x}\sim p_{\theta}(\mathbf{x}\mid\mathbf{z})$.


VAEs offer a continuous latent space that supports interpolation, sampling, and, in conditional variants, controllable generation and editing. However, standard VAEs often produce overly smooth samples and can suffer from posterior collapse. Although more expressive priors and approximate posteriors~\cite{rezende2015flows,kingma2016iaf} can improve results, VAEs generally fall short of GANs~\cite{goodfellow2014gan} and diffusion models~\cite{rombach2022ldm} in perceptual fidelity.


\subsubsection{Generative Adversarial Networks}
\label{sssec:gans}

\begin{figure}[t]
  \centering
  \includegraphics[width=0.99\linewidth]{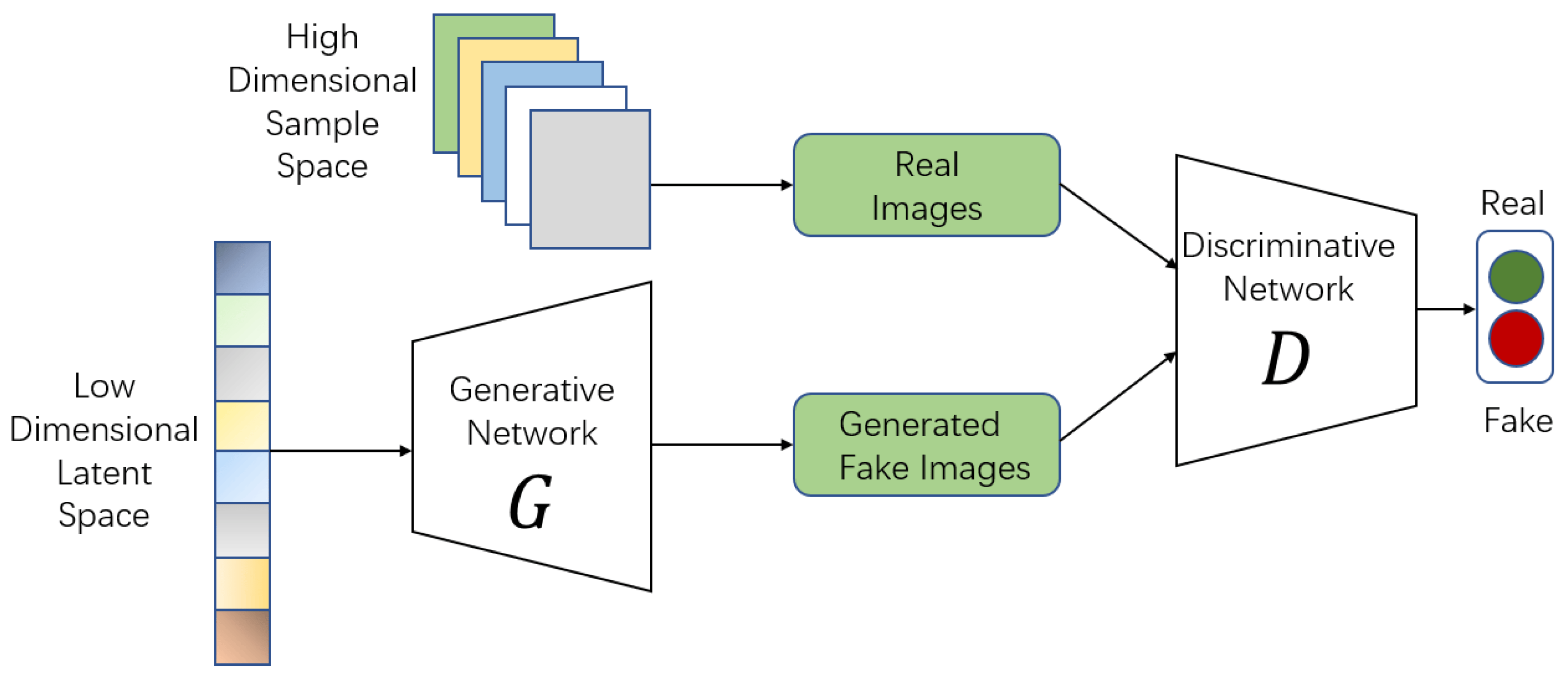}
  \caption{
  %
  An overview of generative adversarial networks (GANs). A generator $G$ maps noise samples (latent codes) to synthetic images, while a discriminator $D$ learns to distinguish real images from generated ones; both are trained adversarially using $\mathcal{L}_{\text{GAN}}$ (see Sec.~\ref{sssec:gans}). Figure reproduced from~\cite{cai2020ganimage}.
  }
  \label{fig:gan}
\end{figure}

Generative adversarial networks (GANs) formulate generative modeling as a two-player game between a generator $G$ and a discriminator $D$~\cite{goodfellow2014gan}. The classical minimax objective
\[
\mathcal{L}_{\text{GAN}} = \min_{G}\max_{D}\; \left[\mathbb{E}_{\mathbf{x}\sim p_{\text{data}}}[\log D(\mathbf{x})]
+ \mathbb{E}_{\mathbf{z}\sim p(\mathbf{z})}[\log (1-D(G(\mathbf{z})))]\right]
\]
trains $D$ to distinguish real samples $\mathbf{x}$ from synthesized ones $G(\mathbf{z})$, while $G$ learns to fool $D$ (Fig.~\ref{fig:gan}). Under an optimal discriminator, this objective can be shown to correspond to minimizing the Jensen--Shannon divergence between the data and model distributions, which motivated later variants based on alternative divergence measures.

In fact, numerous extensions have strengthened the original GAN formulation, improving training stability, sample fidelity, and controllability. For instance, Wasserstein GANs~\cite{arjovsky2017wgan,gulrajani2017wgangp} reformulate adversarial training using the Wasserstein-1 (Earth-Mover) distance, yielding smoother gradients and more stable optimization. Complementarily, PatchGAN discriminators classify local image patches rather than entire images, enabling sharper detail in conditional settings such as image-to-image translation~\cite{isola2017pix2pix}.

However, common challenges in training GANs include instability (\textit{e.g.}, oscillations or failure to converge), mode collapse (loss of diversity), and sensitivity to the choice of objective and regularization. While Wasserstein objectives and gradient penalties help mitigate these issues, they do not eliminate them entirely~\cite{arjovsky2017wgan,gulrajani2017wgangp}. Nonetheless, GANs remain attractive for high-fidelity, sharp synthesis and for learning class- or image-conditional mappings with controllable outputs.


\subsubsection{Diffusion Models}

Diffusion models learn to reverse a process that gradually perturbs data into Gaussian noise~\cite{sohl2015deep,ho2020ddpm}. The forward process is defined by a variance schedule $\{\beta_t\}_{t=1}^{T}$:
\begin{equation*}
\begin{gathered}
q(\mathbf{x}_t\mid \mathbf{x}_{t-1})=\mathcal{N}\!\big(\sqrt{1-\beta_t}\,\mathbf{x}_{t-1},\,\beta_t\mathbf{I}\big), \\
\bar{\alpha}_t=\textstyle\prod_{s=1}^t(1-\beta_s), \quad
\mathbf{x}_t=\sqrt{\bar{\alpha}_t}\,\mathbf{x}_0+\sqrt{1-\bar{\alpha}_t}\,\boldsymbol{\epsilon},\ \boldsymbol{\epsilon}\!\sim\!\mathcal{N}(\mathbf{0},\mathbf{I}).
\end{gathered}
\end{equation*}
A neural network $\epsilon_{\theta}$ is then trained to predict the injected noise from $(\mathbf{x}_t,t)$~\cite{ho2020ddpm}. In the commonly used simplified objective, this becomes
\[
\mathcal{L}_{\mathrm{DDPM}}=\mathbb{E}_{\mathbf{x}_0,t,\boldsymbol{\epsilon}}\!\left[\big\|\boldsymbol{\epsilon}-\epsilon_{\theta}(\mathbf{x}_t,t,\mathbf{c})\big\|_2^2\right].
\]
Conditioning variables $\mathbf{c}$ (\textit{e.g.}, text features) can be incorporated in several ways, including guidance at sampling time and conditioning mechanisms within the denoiser such as cross-attention.

\begin{figure}[t]
  \centering
  \includegraphics[width=0.99\linewidth]{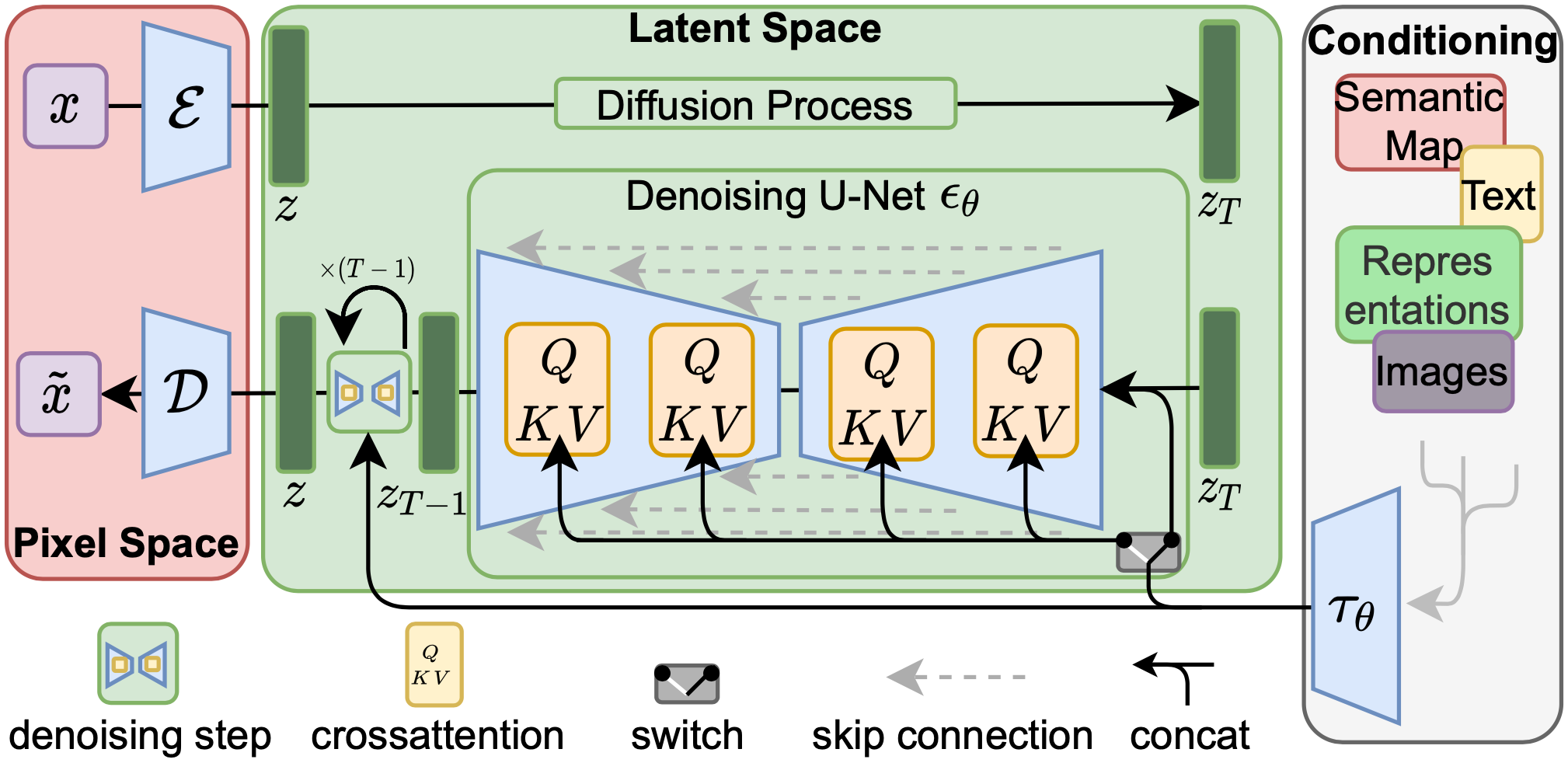}
  \caption{
  An overview of latent diffusion models (LDMs). An autoencoder $(\mathcal{E},\mathcal{D})$ encodes images $\mathbf{x}$ into latents $\mathbf{z}$ (and decodes back to $\tilde{\mathbf{x}}$), and a denoising network $\epsilon_{\theta}$ iteratively predicts and removes noise; conditioning (\textit{e.g.}, text/images/semantic maps) is injected via cross-attention. Figure reproduced from~\cite{rombach2022ldm}.
  }
  \label{fig:dm_ldm}
\end{figure}

Direct diffusion in pixel space is computationally expensive. Latent diffusion models mitigate this by learning an autoencoder $(\mathcal{E},\mathcal{D})$ and performing the diffusion process in a lower-dimensional latent space $\mathbf{z}=\mathcal{E}(\mathbf{x})$, with conditioning (\textit{e.g.}, text) injected via cross-attention inside the denoiser~\cite{rombach2022ldm} \revone{(Fig.~\ref{fig:dm_ldm})}. Beyond text prompts, additional conditioning modules have been introduced to provide finer control without retraining the full backbone. ControlNet~\cite{zhang2023controlnet} adds condition-specific, zero-initialized branches to incorporate structural signals such as edges, depth, or human pose while preserving the pretrained model. IP-Adapter~\cite{ye2023ipadapter} introduces a lightweight image-prompt pathway through decoupled cross-attention, enabling style and identity control alongside text conditioning.


Diffusion has been applied to a wide range of modalities, including images~\cite{rombach2022ldm}, video~\cite{ho2022vdm}, audio~\cite{liu2023audioldm}, and 3D content~\cite{lin2023magic3d,chen2023fantasia3d}, as well as mesh-based textures, as discussed in subsequent sections. Collectively, the core formulation, latent variants, and conditioning mechanisms have evolved into a flexible toolkit that has broadened the practicality of diffusion-based generation.



\section{Guidance for 3D Mesh Texturing}
\label{sec:guidance}

This section discusses the types of \emph{guidance} used in neural 3D mesh texturing. Here, guidance refers to the signals or conditions that specify aspects of the desired texture and steer the generative model toward a target result. Broadly, we group guidance into two categories: \textit{structural} guidance, which constrains the spatial layout or geometric placement of texture on the 3D surface, and \textit{stylistic} guidance, which specifies the desired appearance of the texture.


\subsection{Structural Guidance}

\begin{figure}[t]
  \centering
  \includegraphics[width=0.75\linewidth]{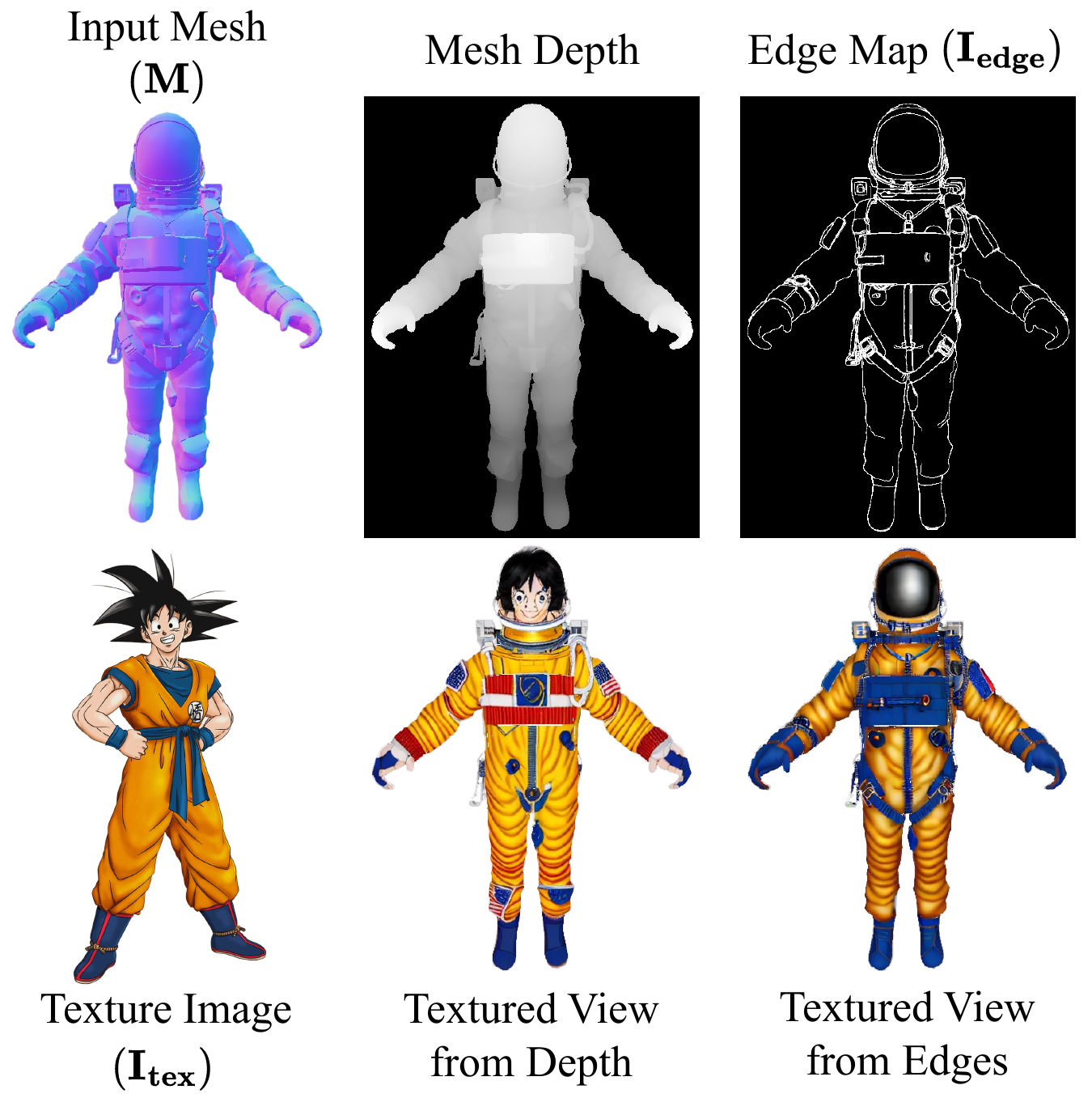}
  \caption{
  %
  Depth vs.\ edges for structural guidance. Depth maps are typically smooth and provide coarse structural cues. In contrast, geometric edges derived from mesh attributes (\textit{e.g.}, normals, depth discontinuities, and connectivity) are more detailed and aligned with mesh geometry, improving mesh--texture consistency. Figure reproduced from~\cite{perla2024easitex}.
  }
  \label{fig:depth_v_edges}
\end{figure}

Structural guidance provides geometric cues that inform the model about \textit{where} specific content should appear on the mesh. It does not define the artistic style of those elements, but helps ensure that generated textures remain consistent with the underlying geometry and part layout of the mesh. Because it is tied to the 3D surface, such guidance is often derived from the input mesh or from mesh-aligned intermediate representations. Regardless of model design or stylistic conditioning, many methods condition the generation process on structural information to avoid textures that conflict with geometry. Several forms of structural guidance have been explored, including geometric edges derived from mesh attributes (normals, depth, connectivity)~\cite{perla2024easitex} (Fig.~\ref{fig:depth_v_edges}), depth maps~\cite{richardson2023texture,chen2023text2tex}, normal maps~\cite{yeh2024texturedreamer}, canonical UV layouts of SMPL/SMPL-X human templates~\cite{loper2015smpl,pavlakos2019smplx,liu2024texdreamer}, and silhouettes~\cite{yu2021learning_texture_generators}. More generally, any representation that provides a correspondence between 3D surface regions and 2D texture locations can serve as structural guidance. The choice and integration of such cues usually depend on the task and model design.

\subsection{Stylistic Guidance}

Stylistic guidance defines \textit{what} the texture should look like---its color palette, materials, patterns, or overall appearance---independent of geometry. It specifies the desired appearance of the texture and is typically provided through an external conditioning signal, such as text, reference images, or exemplar textures. Texture generation can also proceed without such input, \textit{i.e.}, unconditionally, in which case the style of the output is drawn solely from the learned training distribution. Conversely, when conditioning is provided, the model is guided to match the specified stylistic attributes.

\subsubsection{Unconditional Texture Generation}
\label{subsec:uncond_gen}

\begin{figure}[t]
  \centering
  \includegraphics[width=0.99\linewidth]{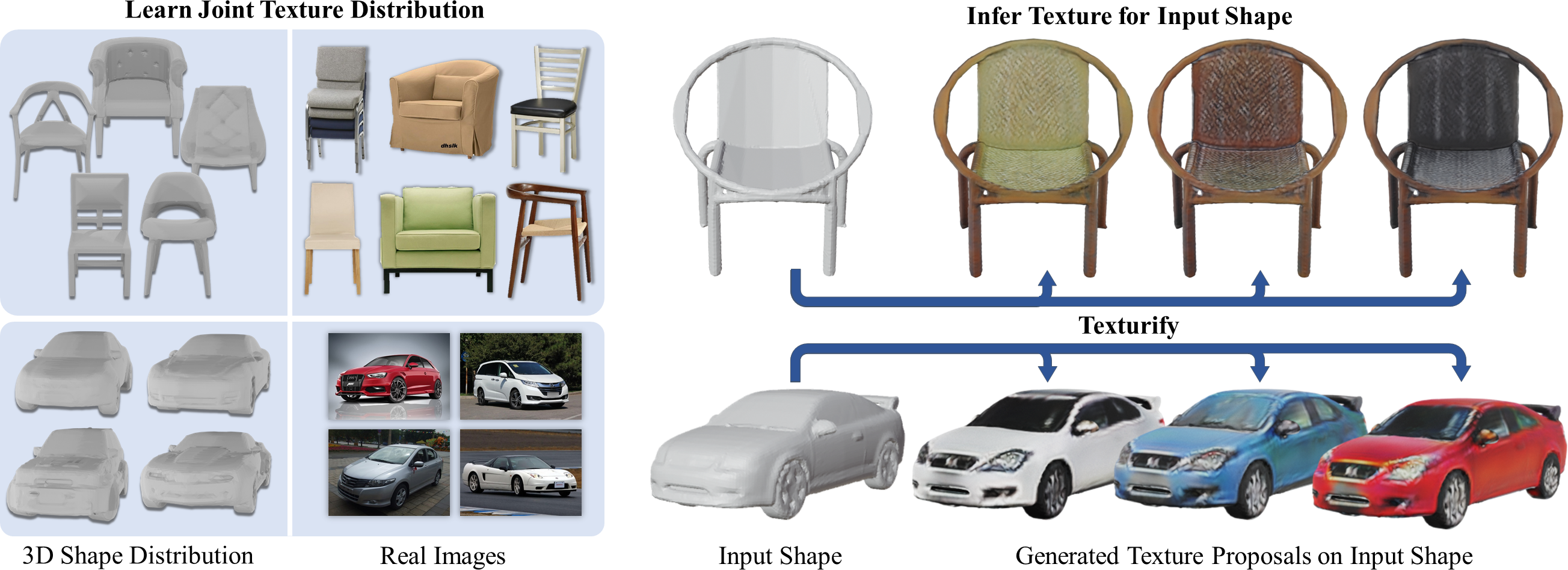}
  \caption{
  Unconditional texture generation with Texturify. The model learns a distribution of plausible textures conditioned on 3D shape from real-image supervision, and infers diverse texture proposals for a given input shape. Figure reproduced from~\cite{siddiqui2022texturify}.
  }
  \label{fig:uncond_gen}
\end{figure}

Unconditional texture generation produces textures without explicit stylistic guidance, instead sampling appearance from a learned distribution, typically from random noise or a latent code. In this setting, the model may still be conditioned on the mesh or other structural inputs, but it is not given user-specified appearance cues. Without such high-level conditioning, specific attributes cannot be directly controlled; however, varying the random seed or latent input allows diverse sampling from the learned distribution, with limited control possible through latent-space manipulations, as in 2D GANs~\cite{karras2019stylegan,karras2020stylegan2} (Fig.~\ref{fig:uncond_gen}).


The main advantage of unconditional generation is its simplicity and ability to produce diverse textures without user input. However, the absence of explicit appearance control is a major limitation, making it less practical when a specific target style is required. On the positive side, unconditional generation is often relatively straightforward to integrate, since randomness can be introduced through latent or noise inputs without requiring additional conditioning signals.

\subsubsection{Conditional Texture Generation}
\label{subsec:cond_gen}

Conditional texture generation provides explicit control over the style or appearance of the output through additional input cues. The model receives external conditioning signals that specify the desired texture and uses them to guide generation. Stylistic guidance can take various forms---such as text, reference images, exemplar textures, or textured 3D assets. Conditioning greatly expands user control and broadens the range of possible tasks (\textit{e.g.}, texture transfer, domain-specific texturing, and style interpolation), but also introduces challenges: the model must faithfully reproduce the specified style while maintaining coherence with the underlying mesh structure. The choice of conditioning modality often influences the model architecture, training strategy, and target applications.


Conditional texture generation is often evaluated along two dimensions: \textit{(i)} the intrinsic quality and realism of the generated texture, and \textit{(ii)} its faithfulness to the input guidance. High-quality textures that deviate from the prompt, or prompt-faithful results that lack realism, are both inadequate; achieving both plausibility and guidance adherence remains a central goal of conditional methods.

\paragraph*{Text-based Generation.}

\begin{figure}[t]
  \centering
  \includegraphics[width=0.99\linewidth]{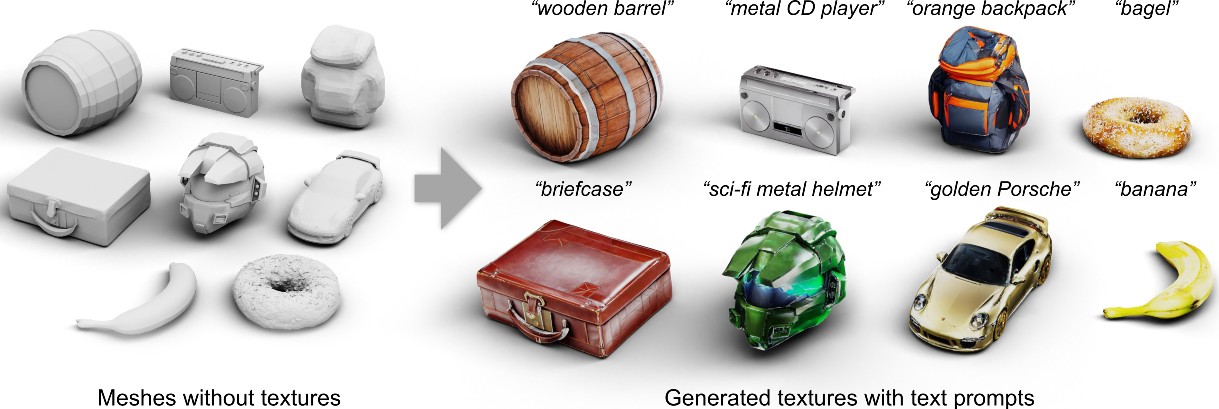}
  \caption{Text-conditioned texture generation. Given untextured meshes (left), Text2Tex synthesizes textures guided by text prompts (right). Figure reproduced from~\cite{chen2023text2tex}.
  }
  \label{fig:cond_gen}
\end{figure}

Using natural language as guidance is one of the most popular and convenient ways to control 3D texture synthesis, owing to the accessibility of text as an interface. In text-guided texturing, a user provides a description (prompt) of the desired texture, and the model conditions generation on this input to produce a semantically matching result (Fig.~\ref{fig:cond_gen}). Text is typically incorporated via a pretrained encoder that converts the prompt into a conditioning representation, which is then injected into the generation pipeline. Many systems use vision--language models such as CLIP~\cite{radford2021clip} or dedicated text encoders like T5~\cite{raffel2020t5}, which are usually pretrained on large text or image--text datasets and may either be used as is or fine-tuned for the task at hand.

Text guidance provides high-level, intuitive control, allowing users to specify abstract concepts and attributes that are otherwise difficult to quantify. Its drawback is ambiguity; text can be underspecified or open to multiple interpretations. Diffusion-based methods can help mitigate this, as large text-to-image models like Stable Diffusion~\cite{rombach2022ldm} are trained on vast datasets and can translate rich textual cues into detailed imagery. Leveraging such models allows text-guided 3D texturing to inherit this semantic knowledge.

\paragraph*{Image-based Generation.}

Using an image example as guidance provides a more concrete and precise specification of texture style. Instead of describing the appearance in words, the user provides one or more reference images, and the model generates a texture that matches their visual characteristics. Such guidance can capture details that are difficult to express textually, \textit{e.g.}, specific patterns, artistic styles, or complex color gradients. In image-based texture generation, the reference image is typically encoded into a conditioning representation using a visual encoder~\cite{radford2021clip,simonyan2015vgg}, or incorporated through image-conditioning modules~\cite{ye2023ipadapter}, which then guide texture generation toward the desired style/appearance.

The advantage of image guidance lies in its specificity, meaning that it is generally less ambiguous than text. A reference image can convey fine texture details (for instance, the floral pattern on a dress or the wood grain of a chair) that would be tedious to describe verbally, making this a highly practical form of guidance.

\paragraph*{Textured 3D Shape-based Generation.}

A more structured form of stylistic guidance uses a textured 3D shape, or a mesh-aligned appearance representation, to guide the texturing of another mesh. In this setting, appearance is specified directly in 3D or in an aligned surface domain, rather than through a text prompt or a single reference image. Such guidance is particularly useful for tasks such as texture alignment and transfer across related shapes~\cite{chen2022auvnet}, as well as generating variations of an existing textured asset and, in some cases, transferring learned appearance statistics to new geometries~\cite{mitchel2024fieldlatents}.

Overall, using textured 3D shapes as guidance is less common than text or image inputs, mainly because such exemplars are relatively scarce and often task-specific. When available, however, they can provide detailed and geometrically aligned appearance information. Future systems may combine multiple forms of guidance, \textit{e.g.}, text prompts, reference images, and partial or complete 3D textures, to provide users with finer control over 3D texture creation.

\section{Neural 3D Mesh Texturing}
\label{sec:neural-texturing}

In this section, we review works on \textit{Neural 3D Mesh Texturing} and organize them according to recurring methodological families, while also reflecting the field’s evolution over time. We begin with foundational neural mesh texturing methods that laid the groundwork for later approaches by using differentiable rendering, weak 2D supervision, and adversarial learning to generate textures from limited 2D supervision. We then discuss optimization-based methods, which iteratively refine textures using pretrained priors such as vision--language or diffusion models. Finally, we review accelerated diffusion-based methods, which can be further categorized by inference-time strategy into iterative view-by-view pipelines, synchronized multi-view approaches, and feed-forward methods. This taxonomy reflects a trade-off space between quality, speed, and controllability, and highlights the evolution from slower but flexible optimization to more scalable generation while maintaining consistency across the 3D surface. We provide a summary of representative works in \textit{Neural 3D Mesh Texturing} in Tab.~\ref{tab:all}.



\subsection{Foundational Neural Mesh Texturing}
\label{sec:early-texturing}

Early works on neural 3D mesh texturing demonstrated that neural networks can learn to synthesize realistic textures on 3D surfaces or in surface-aligned representations from limited or indirect supervision. Many of these methods rely on differentiable rendering to bridge the 2D--3D gap, often leveraging unpaired or weakly paired 2D images to learn how to assign realistic textures to 3D meshes. Broadly, these approaches can be grouped into image-driven texturing, texture super-resolution and completion, neural texture representations in function space, and adversarial or GAN-based synthesis pipelines.

\paragraph*{Image-driven Texture Transfer and Reconstruction.}
\emph{PhotoShape}~\cite{park2018photoshape} is an early large-scale effort in this direction, automatically assigning photorealistic appearance to collections of untextured 3D shapes by mining product photographs and material exemplars. Its system retrieves images and materials with appearance similar to a target mesh and aligns them to the shape, enabling large-scale creation of textured ShapeNet-style~\cite{chang2015shapenet} assets. Around the same time, Kanazawa~\etal~\cite{kanazawa2018learning} introduced a category-specific mesh predictor that jointly infers 3D geometry and a corresponding UV texture map from a single image via differentiable rendering, without relying on ground-truth 3D supervision. The model deforms a fixed template mesh with a shared UV atlas to match each target’s geometry while preserving consistent UV coordinates. This canonical UV space gives texels a more stable semantic correspondence across instances, helping disentangle texture from shape and pose. As a result, the network can learn category-level appearance priors efficiently, using the shared template as a strong inductive bias for texture prediction. Building on this idea of category-level canonicalization, Henderson~\etal~\cite{henderson2020leveraging} proposed one of the first fully generative models that learns both shape and texture directly from collections of unpaired 2D images. Unlike Kanazawa~\etal~\cite{kanazawa2018learning}, they omit UV maps entirely, representing texture as piecewise-constant per-face colors on a fixed-topology mesh while predicting vertex positions for geometry. This yields class-consistent face correspondences that enable joint sampling of plausible geometry and appearance from the image distribution, demonstrating that coherent textured shapes can emerge from 2D supervision alone. These methods collectively established that realistic texture generation can be achieved by learning to invert the rendering process using only 2D supervision.

\paragraph*{Texture Super-Resolution and Completion.}
Another challenge for early neural texturing pipelines was the limited resolution of reconstructed textures. Richard~\etal~\cite{richard2019multi} and Li~\etal~\cite{li20193d} addressed this problem through learned texture upsampling and refinement networks. Their CNN-based models fuse multi-view imagery and low-resolution atlases to synthesize sharper, high-frequency texture details that conventional texture-fusion pipelines often fail to recover. In parallel, Chibane and Pons-Moll~\cite{chibane2020implicit} proposed to complete missing texture regions using an implicit representation: their Implicit Feature Network (IF-Net) predicts per-point texture values conditioned on partial scans, enabling plausible inpainting of unseen surfaces consistent with the geometry. These works treat texture refinement as a learnable process, moving beyond simple interpolation (\textit{e.g.}, view-dependent texture sampling~\cite{debevec1996vdtm}) or seam-hiding blends and global color optimization~\cite{zhou2014colormap,waechter2014lettherebecolor,perez2003poisson}.

\begin{figure}[t]
  \centering
  \includegraphics[width=0.99\linewidth]{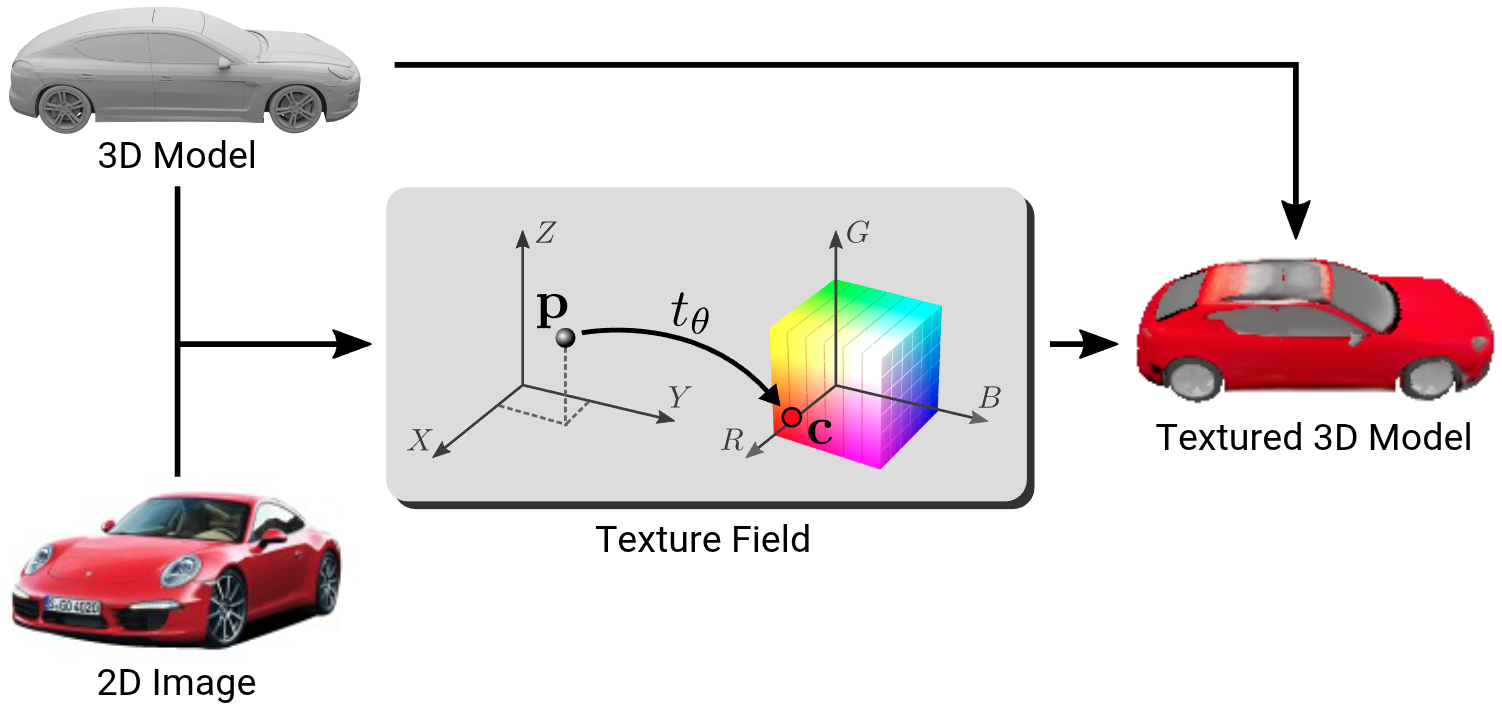}
  \caption{
  An overview of TextureFields. Given a 3D shape (and optionally a reference image), the method learns a continuous texture field $t_\theta$ mapping surface points $\mathbf{p}$ to colors $\mathbf{c}$, producing a textured mesh. Figure reproduced from~\cite{oechsle2019texturefields}.
  }
  \label{fig:texturefields_teaser}
\end{figure}

\paragraph*{Neural Texture Representations in Function Space.}
Oechsle~\etal~\cite{oechsle2019texturefields} proposed \emph{Texture Fields} to represent texture as a continuous neural function that maps spatial queries to RGB color \revone{(Fig.~\ref{fig:texturefields_teaser})}. This implicit formulation reduces dependence on fixed UV atlases and allows resolution-independent texture definition. Texture Fields can be trained jointly with shape representations and, in generative settings, combined with adversarial training~\cite{goodfellow2014gan} to produce high-quality renderings. Recent successors have extended this idea toward neural material representations. For example, TexGaussian~\cite{xiong2025texgaussian} proposed an octree-based 3D Gaussian representation for feed-forward PBR material generation, predicting albedo, roughness, and metallic parameters from geometry-consistent 3D features.

\paragraph*{Adversarial and GAN-based Texture Synthesis.}
Generative Adversarial Networks (GANs) became a prominent direction in early neural texturing pipelines. Huang~\etal~\cite{huang2020adversarial} introduced an adversarial optimization framework for RGB-D scans, refining noisy vertex colors by enforcing photorealism through a learned patch discriminator. By jointly optimizing texture and geometry under differentiable rendering, they achieved sharper and more consistent results than traditional photometric blending. Yu~\etal~\cite{yu2021learning_texture_generators} learned texture generators for ShapeNet~\cite{chang2015shapenet} object collections using only untextured meshes and unaligned internet photos, relying on a common UV layout per class and adversarial 2D supervision. Their network learns to populate the UV map such that renders of the textured meshes match real photographs, establishing an early class-level neural texture generation framework. 3DStyleNet~\cite{yin20213dstylenet} disentangles geometry and texture style in latent space, allowing independent control of shape and appearance and enabling style transfer across objects. SPSG~\cite{dai2021spsg} extends these ideas to full indoor RGB-D scenes through a self-supervised approach that hallucinates missing geometry and surface color for large 3D scans, demonstrating neural texturing at the scene level.

\begin{figure}[t]
  \centering
  \includegraphics[width=0.99\linewidth]{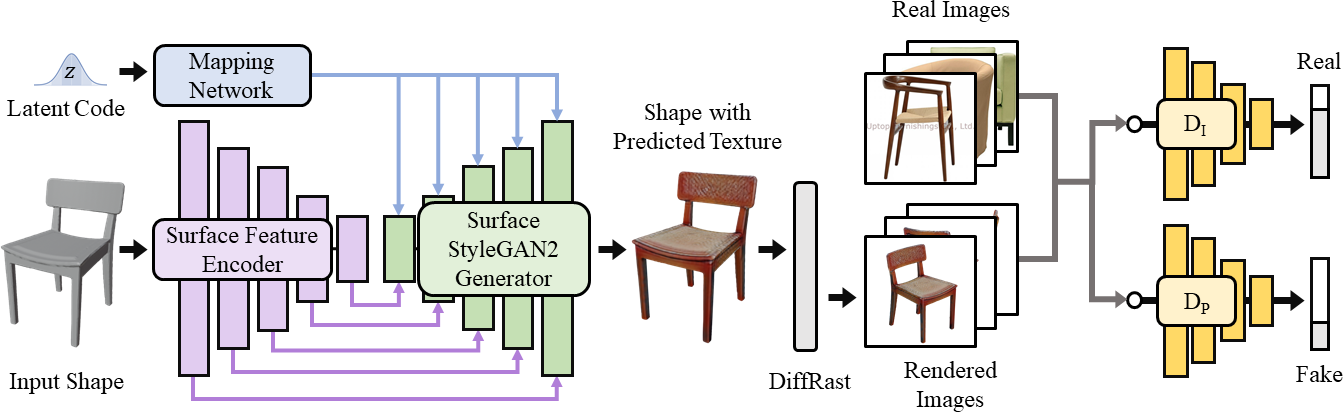}
  \caption{
  Training pipeline of Texturify. Given an untextured input mesh, a surface-feature encoder and a StyleGAN2-based~\cite{karras2020stylegan2} generator predict a texture for the mesh. The textured mesh is then differentiably rendered, and the resulting images are evaluated by two discriminators (a global discriminator $D_I$ and a patch-based discriminator $D_P$) against real images; the adversarial loss trains all networks. Figure reproduced from~\cite{siddiqui2022texturify}.
  }
  \label{fig:texturify_arch}
\end{figure}

\paragraph*{Direct Surface Texture Generation.}
Later works shifted toward generating textures directly on mesh surfaces without relying on explicit UV alignment. \emph{Texturify}~\cite{siddiqui2022texturify} is a surface-based GAN that predicts per-face colors directly on the mesh (Fig.~\ref{fig:texturify_arch}). It leverages a hierarchical 4-RoSy parameterization~\cite{huang2019texturenet} to define orientation-consistent face convolutions, combining a mesh-face encoder with a StyleGAN2-inspired~\cite{karras2020stylegan2} decoder. Training uses multi-view differentiable rendering from random cameras and two discriminators: an image discriminator for overall realism and a patch-consistency discriminator to enforce cross-view consistency. This framework avoids explicit UV atlases and the associated seam/distortion issues, learns geometry-aware textures from unpaired 2D images and untextured meshes, and captures high-frequency detail, with its effective resolution primarily limited by the number of faces at the finest 4-RoSy level.

Mesh2Tex~\cite{bokhovkin2023mesh2tex} extends this line of work by replacing direct per-face color prediction with a \emph{hybrid mesh--neural-field} representation: a face-convolution encoder--decoder produces coarse per-face features, while a shared neural field maps these features to RGB. This formulation enables high-resolution texture generation beyond coarse face-wise color prediction. Mesh2Tex also supports \emph{image-guided} texturing through inference-time optimization, aligning renders of the target mesh with a reference image in a perceptual sense.

\emph{ShaDDR}~\cite{chen2023shaddr} developed an example-based deep generative framework that, given a coarse voxel shape, detailizes geometry and generates textures in a style learned from a small set of textured exemplars. For texture synthesis, differentiable rendering compares multi-view renders of the generated shape against exemplar texture images, while style is controlled through learned latent codes. These methods mark a transition from 2D-conditioned texturing toward more direct neural texture generation on 3D surfaces.

While these neural texturing approaches demonstrated that realistic and controllable 3D texture generation is feasible, they also faced important constraints. Their reliance on limited object categories and small datasets restricted generalization (\textit{e.g.}, category-specific canonical UVs or class-specific shape collections)~\cite{kanazawa2018learning,yu2021learning_texture_generators}, and GAN-based objectives often suffered from instability or limited diversity relative to later generative models~\cite{lucic2018aregansequal}. Many architectures also required meshes with specific structural or parameterization assumptions (\textit{e.g.}, 4-RoSy-based surface representations or class-aligned UV layouts), limiting applicability to arbitrary models~\cite{siddiqui2022texturify,yu2021learning_texture_generators}. Furthermore, text-based or open-vocabulary control was essentially absent in this era---capabilities that only emerged with subsequent vision--language and diffusion frameworks~\cite{chen2022tango,richardson2023texture,chen2023text2tex}. Even methods that achieved high realism, such as \emph{Texturify}~\cite{siddiqui2022texturify} and \emph{Mesh2Tex}~\cite{bokhovkin2023mesh2tex}, remained limited by the fidelity/diversity trade-offs of pre-diffusion generative priors and lacked the broader semantic control and robustness introduced by these newer paradigms. Nonetheless, these pioneering works established the foundations of modern neural texturing by demonstrating that neural networks, coupled with differentiable rendering, can learn to synthesize textures directly on 3D meshes from weak 2D supervision~\cite{kato2018nmr,siddiqui2022texturify,kanazawa2018learning}.

\subsection{Optimization-based Texturing}
\label{subsecmain:opt}

In \emph{optimization-based texturing}, a mesh’s texture (and sometimes geometry) is directly refined to satisfy objectives derived from powerful pre-trained models, leveraging their prior knowledge to generate diverse, detailed textures, often without task-specific training. Such methods offer several benefits, including improved texture consistency: optimizing a shared texture map across many rendered views can help mitigate discontinuities that often affect multi-view feed-forward approaches.

\paragraph*{Vision--Language Model Guidance.}
A prominent optimization-based line of work uses vision--language models such as CLIP~\cite{radford2021clip} to guide texture synthesis. As a precursor, Neural 3D Mesh Renderer~\cite{kato2018nmr} introduced a differentiable renderer and demonstrated that mesh textures (and even vertices) can be optimized using image-based objectives. In particular, it showed that a mesh can be iteratively updated to minimize a 2D \emph{style loss}~\cite{gatys2016styletransfer}, enabling style transfer from an image onto a 3D asset. Building on this idea, later methods replaced style images with natural-language descriptions using CLIP’s joint vision--language embedding. In these approaches, the mesh is rendered from multiple viewpoints, and the CLIP loss between rendered images and the target text is used to update the texture. For instance, Text2Mesh~\cite{michel2022text2mesh} optimizes mesh appearance and local geometric detail to match a given prompt in CLIP space, enabling compelling text-driven stylization through optimization alone. Subsequent works improved efficiency and fidelity: X-Mesh~\cite{ma2023xmesh} introduced a text-guided dynamic attention mechanism that improves stylization accuracy and convergence speed while refining both geometry and texture. When geometry deformation is disabled, the method reduces to a purely text-guided texture optimization pipeline.

Several related methods also use CLIP-based objectives for text-driven 3D appearance optimization~\cite{khalid2022clipmesh,jetchev2021clipmatrix,jain2022dreamfields}. While CLIP-guided texture optimization can produce broadly plausible and semantically relevant results without task-specific 3D training data, its visual quality is often limited by the coarse supervision provided by CLIP embeddings. To push toward photorealism, TANGO~\cite{chen2022tango} extends this line of work to optimize spatially varying material properties, local geometric variation, and lighting under a differentiable renderer. By predicting material, normal map, and lighting condition from a text prompt, TANGO produces more realistic appearances, including shiny or metallic finishes, and enables photorealistic text-driven stylization. Overall, CLIP-based methods established the feasibility of text-driven texturing and stylization, but were later surpassed in fidelity by diffusion-based approaches.

\paragraph*{Diffusion-Driven Optimization.}
Instead of relying on CLIP embedding alignment, some methods optimize textured meshes using guidance from a pre-trained diffusion model, \textit{e.g.}, Stable Diffusion~\cite{rombach2022ldm}, conditioned on the same text prompt. The key idea, introduced by DreamFusion~\cite{poole2023dreamfusion}, is \emph{Score Distillation Sampling} (SDS): rather than comparing against a single target image, SDS uses the score function of a frozen text-to-image diffusion model to provide gradients that drive the 3D representation toward the prompt-conditioned image distribution. In practice, this is often written as an SDS objective $\mathcal{L}_{\text{SDS}}$ whose gradients encourage rendered views of the mesh to be interpreted by the diffusion model as matching the prompt. Optimizing this objective directly on a mesh’s texture can synthesize more complex and higher-frequency details than earlier CLIP-based methods.

A limitation of basic SDS, however, is oversmoothing or oversaturation of textures, partly due to how the guidance gradients are obtained. To address this, ProlificDreamer~\cite{wang2023prolificdreamer} proposed \emph{Variational Score Distillation} (VSD). VSD treats the desired 3D texture as a random variable and optimizes a variational bound, which introduces diversity and reduces bias in score estimation. As a result, VSD can yield sharper and more diverse details than basic SDS. Although DreamFusion and ProlificDreamer optimize NeRF-based representations, the same principles can transfer to explicit mesh textures. In fact, \emph{Latent-NeRF}~\cite{metzer2023latentnerf} showed that latent-space score distillation can be applied directly to a mesh’s UV texture map. In their scheme, the texture is represented in the latent space of a pretrained autoencoder; SDS is applied in that compact space to generate a coarse latent texture, which is then decoded to RGB and further refined. This yields high-resolution textures more efficiently than pixel-space optimization.

\begin{figure}[t]
  \centering
  \includegraphics[width=0.99\linewidth]{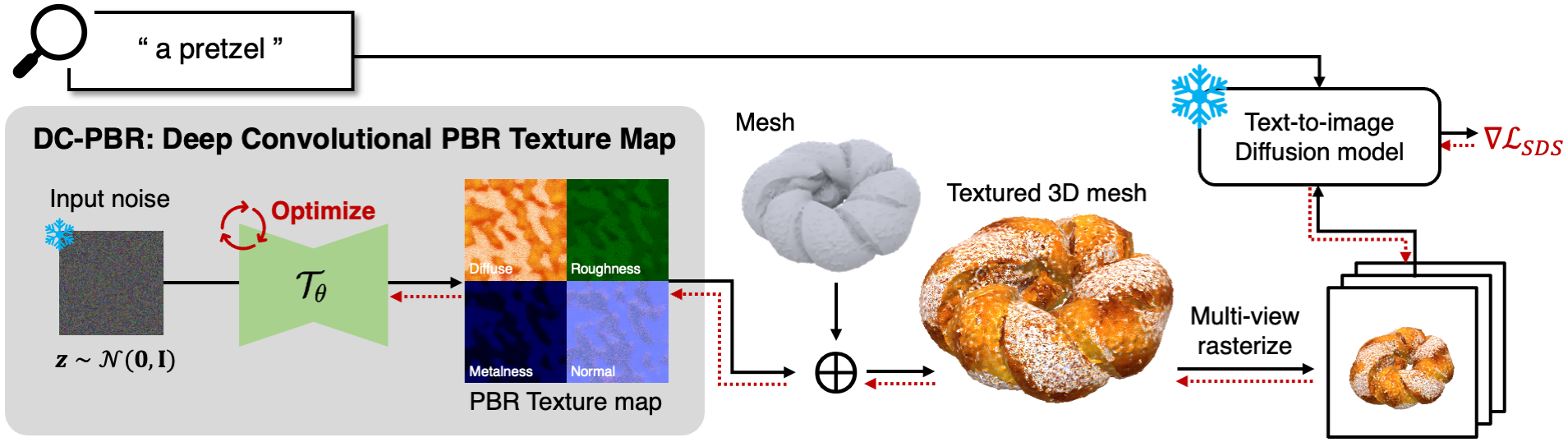}
  \caption{
  An overview of Paint-It. Given an untextured mesh and a text prompt, the method optimizes PBR texture maps (\textit{e.g.}, albedo, roughness, metalness, normals) using SDS guidance~\cite{poole2023dreamfusion} from a frozen text-to-image diffusion model over multi-view renderings. Figure reproduced from~\cite{kim2024paintit}.
  }
  \label{fig:paintit_arch}
\end{figure}

Subsequent works extended this paradigm to physically based material generation. Fantasia3D~\cite{chen2023fantasia3d} demonstrated text-to-3D content creation with disentangled geometry and appearance. By optimizing an explicit mesh for shape and a neural appearance representation for material under SDS guidance, Fantasia3D generates textured meshes with relightable PBR material properties from a text prompt. Building on this direction, Paint-It~\cite{kim2024paintit} targets high-fidelity physically based texture generation \revone{(Fig.~\ref{fig:paintit_arch})}. The authors observed that applying SDS directly to pixel-wise texture maps leads to noisy and blurry results due to uneven gradient coverage. To address this, they represent each texture map (diffuse, specular, roughness) with a lightweight convolutional network instead of raw pixels. This re-parameterization acts as an implicit regularizer, filtering high-frequency noise from diffusion gradients and enabling coarse-to-fine optimization. Paint-It achieves faster convergence and finer details (e.g., text and engravings) than prior distillation-based methods, while also allowing explicit lighting control during texture generation. FlashTex~\cite{deng2024flashtex} introduces LightControlNet, an illumination-aware diffusion model conditioned on user-specified lighting (e.g., HDRI or text). Its pipeline first generates lighting-consistent multi-view images of the mesh and then refines the texture through optimization so that the recovered appearance remains relightable and consistent under arbitrary illumination. Going a step further, DreamMat~\cite{zhang2024dreammat} trains a diffusion model that is explicitly geometry- and light-aware. The model conditions on geometric cues such as normal and depth maps together with lighting and text, and learns to generate physically plausible material maps. Relatedly, DreamPBR~\cite{xin2025dreampbr} targets text-driven, high-resolution SVBRDF generation with multimodal guidance, producing relightable PBR parameter maps that can complement mesh-based material texturing.

More recently, video diffusion models have been explored for texturing because they often provide stronger inter-frame, and thus cross-view, consistency than image-based generators. For example, VideoMat~\cite{munkberg2025videomat} extracts relightable PBR material maps for a given 3D shape by first generating multi-view, view-consistent renderings with a fine-tuned video diffusion model, then recovering base color, roughness, and metallic through intrinsic decomposition, and finally applying differentiable path-traced refinement to output standard PBR maps.


Beyond text prompts, optimization-based texturing has been adapted to other forms of guidance. TextureDreamer~\cite{yeh2024texturedreamer} addresses image-driven texture synthesis. Given only 3--5 reference photos of a real object or scene, it personalizes a diffusion model to capture the reference appearance, in a manner inspired by DreamBooth~\cite{ruiz2023dreambooth}, and then optimizes the target mesh’s relightable texture maps using VSD~\cite{wang2023prolificdreamer} so that rendered views match the references.

In a similar spirit, StyleTex~\cite{xie2024styletex} focuses on style transfer from a single 2D image. It decouples the image’s style from its content in CLIP space, then injects the style features via cross-attention during diffusion-guided, multi-view optimization of the mesh’s UV texture, while using the content features as negative guidance to suppress content leakage. This yields RGB (non-PBR) textures that faithfully inherit the reference image’s artistic style (color palettes, brushstrokes, \emph{etc.}) without distorting the mesh structure. Another intriguing direction is optimizing procedural material parameters instead of raw textures. MaPa~\cite{zhang2024mapa} segments a 3D model and assigns each part a procedural material graph, as used in tools like \textit{Blender}~\cite{blenderManual2025} or \textit{Substance 3D}~\cite{adobeSubstancePainter2025}. The parameters of these graphs (\textit{e.g.}, noise scale, color, roughness) are optimized using a segment-controlled text-to-image diffusion model that synthesizes part-aligned target images, bridging text descriptions to material parameters without paired training data. This produces high-quality, tileable textures with correct reflectance and, crucially, editable procedural materials rather than baked bitmaps.

There have also been efforts to texturize an entire scene. SceneTex~\cite{chen2024scenetex} addresses this challenge for large indoor 3D scenes with many objects from a text prompt. Instead of optimizing a single explicit scene-level UV atlas, it represents appearance using a multi-resolution texture field and optimizes it with a score-distillation-based objective. To ensure global coherence, SceneTex introduces view-dependent refinement through a cross-attention decoder that propagates appearance information across viewpoints, yielding fully textured rooms in which walls, floors, and furniture share a consistent style.

\begin{figure}[t]
  \centering
  \includegraphics[width=0.99\linewidth]{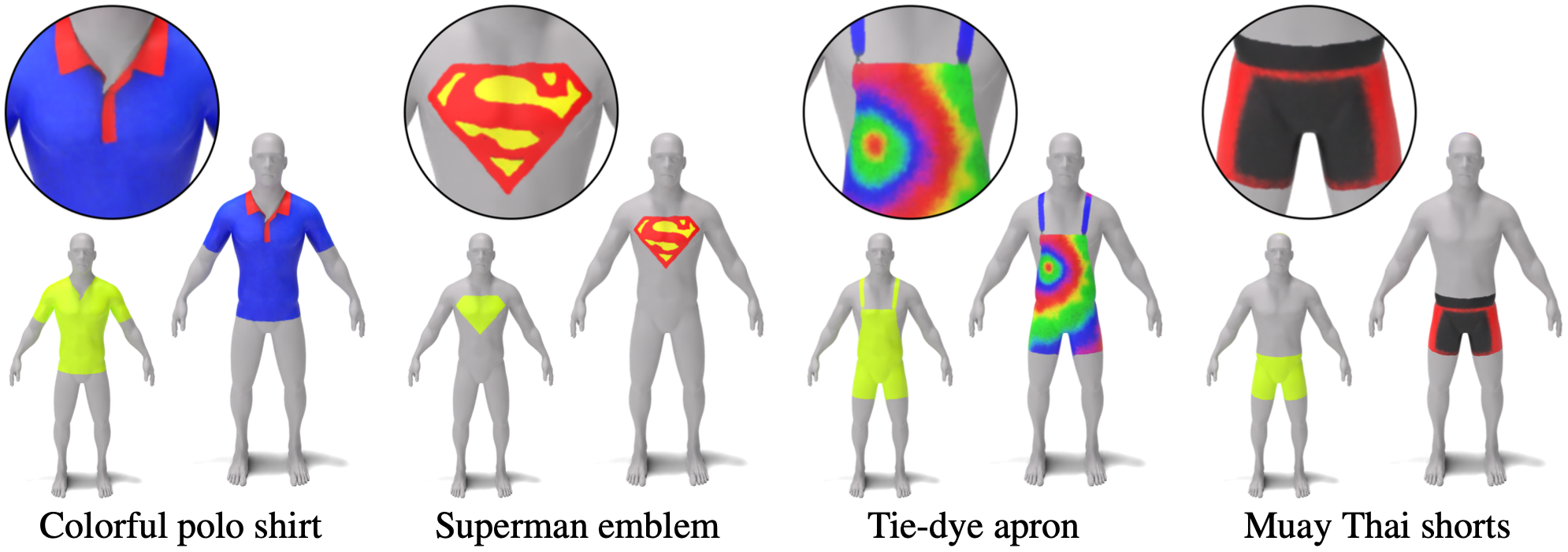}
  \caption{
  Given a text prompt, 3D Paintbrush predicts both the target region and corresponding texture on the input mesh, enabling spatially controlled texturing. Figure adapted from~\cite{decatur2024paintbrush}.
  }
  \label{fig:3d_paintbrush_res}
\end{figure}

Finally, optimization can also be constrained to \emph{local} regions of a mesh. 3D Paintbrush~\cite{decatur2024paintbrush} enables localized text-driven texturing using a novel Cascaded Score Distillation (CSD) loss that combines guidance from multiple diffusion stages. Given a mesh and a text prompt such as \textit{``superman emblem''}, it jointly learns a localization map to identify the target region and a texture map, both represented as neural fields, confining the edit to the predicted area while leaving the rest of the texture unchanged (Fig.~\ref{fig:3d_paintbrush_res}).

Optimization-based texturing has significantly advanced the quality and flexibility of 3D asset creation, but it remains limited by runtime. Unlike feed-forward networks that generate textures in a single pass, these methods typically require many optimization and denoising steps per shape. Even with accelerations such as \emph{latent-space distillation/optimization}~\cite{metzer2023latentnerf} or \emph{efficient differentiable rendering}~\cite{laine2020nvdiffrast,nimier2019mitsuba2,jakob2022drjit,ravi2020pytorch3d}, texturing a single model can still take several minutes, making these approaches costly, especially for time-sensitive or large-scale applications without substantial compute. Nonetheless, optimization-based texturing opened a new frontier: by leveraging powerful 2D priors, it enables automatic texture generation at a level of detail and semantic richness that was previously difficult to achieve. Ongoing research seeks to reduce runtimes through \emph{better initialization and staged pipelines}~\cite{lin2023magic3d}, \emph{parallelism across views}, and \emph{hybrid learning--optimization designs} that combine generative priors with render-and-backprop refinement~\cite{deng2024flashtex,yeh2024texturedreamer}. At the same time, work on \emph{creative control}---\textit{e.g.}, localized edits and procedural/material parameter optimization---has expanded user control over where and how edits are applied~\cite{decatur2024paintbrush,zhang2024mapa,guo2023decorate3d}. In this sense, combining differentiable rendering with powerful vision--language and diffusion priors remains a compelling paradigm for high-quality 3D mesh texturing.

\subsection{Accelerated Diffusion-based Texturing}
\label{subsecmain:diff_paint}

Early works demonstrated that 2D diffusion models can be used to optimize 3D textures (\textit{e.g.}, by score distillation~\cite{poole2023dreamfusion}). While these optimization-based methods achieve impressive results, they come at the cost of lengthy per-object optimization (often tens of minutes). The need for faster, more scalable approaches motivated a new class of methods that avoid costly test-time optimization by instead using diffusion models in one of three ways: (1) iterative view-by-view painting with a frozen 2D diffusion prior, (2) synchronized multi-view generation that diffuses all views concurrently, or (3) feed-forward texture synthesis in which a custom diffusion model directly outputs a full texture map.

\begin{figure}[t]
  \centering
  \includegraphics[width=0.99\linewidth]{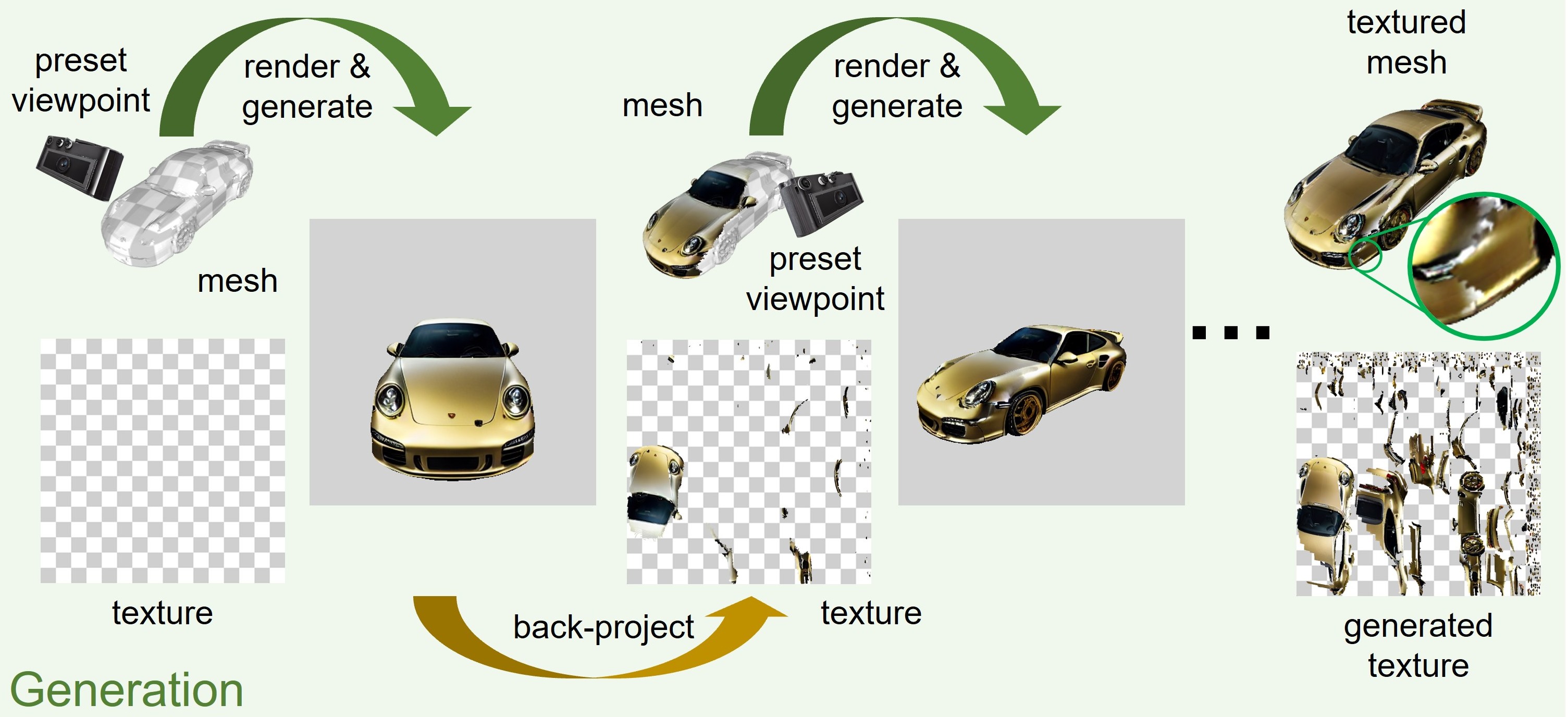}
  \caption{
  An overview of an iterative texturing pipeline. The mesh is rendered from preset viewpoints, a text-to-image model generates view-consistent appearances, and the results are back-projected to update the texture map, iterating across views. Figure adapted from~\cite{chen2023text2tex}.
  }
  \label{fig:text2tex_arch}
\end{figure}

\paragraph*{Iterative Texturing.}
A widely adopted pipeline ``paints'' the mesh one view at a time in a loop: (i) render the current viewpoint with geometric cues (\textit{e.g.}, depth or normals); (ii) use a pretrained image diffusion model to synthesize an image consistent with the current view and prompt; (iii) back-project the synthesized colors onto the UV map using visibility; and (iv) update a per-texel progression mask before moving to the next view~\cite{richardson2023texture,chen2023text2tex}. Dynamic keep/refine/generate masks guide denoising toward unpainted or low-confidence regions while protecting already generated high-quality textures, progressively covering the UV space across multiple views~\cite{richardson2023texture,chen2023text2tex} \revone{(Fig.~\ref{fig:text2tex_arch})}. Many implementations also include a final seam-refinement pass to clean residual discontinuities along chart boundaries~\cite{chen2023text2tex}. In practice, this iterative scheme can yield high-quality textures without fine-tuning the diffusion backbone, although minor seams, color drift, or incomplete coverage may remain across view boundaries~\cite{richardson2023texture,chen2023text2tex}.

Several works build on the iterative painting paradigm for specialized scenarios. EASI-Tex~\cite{perla2024easitex} targets single-image texture transfer: given a reference photo, it uses an IP-Adapter~\cite{ye2023ipadapter} to diffuse the photo’s texture onto a 3D shape. It also showed that geometric edges yield more structurally accurate textures than depth or normals. Paint3D~\cite{zeng2024paint3d} follows a coarse-to-fine strategy: it first samples a pretrained diffusion model to obtain a coarse texture, then uses custom UV-space diffusion models to upsample, inpaint missing regions, and remove baked-in lighting. In the interest of speed, Make-A-Texture~\cite{xiang2025make} optimizes the diffusion model and introduces a specialized backprojection algorithm that generates a full texture in only 3 seconds, trading some detail for significant speed gains. Iterative pipelines have also been adapted to complex scenes: InstanceTex~\cite{yang2024instancetex} textures multi-object scenes instance by instance while aiming to maintain global coherence, whereas RoomTex~\cite{wang2024roomtex} unwraps indoor scenes into panoramas for an initial global pass, then iteratively inpaints each object with panoramic guidance to mitigate inter-object style inconsistencies. Despite these advances, iterative methods can still exhibit minor seams or blur due to view-wise generation, though they remain far faster and more practical than per-mesh optimization, requiring a fixed number of diffusion denoising steps per view instead of long gradient-based optimization.

\paragraph*{Synchronized Texturing.} 

\begin{figure}[t]
  \centering
  \includegraphics[width=0.99\linewidth]{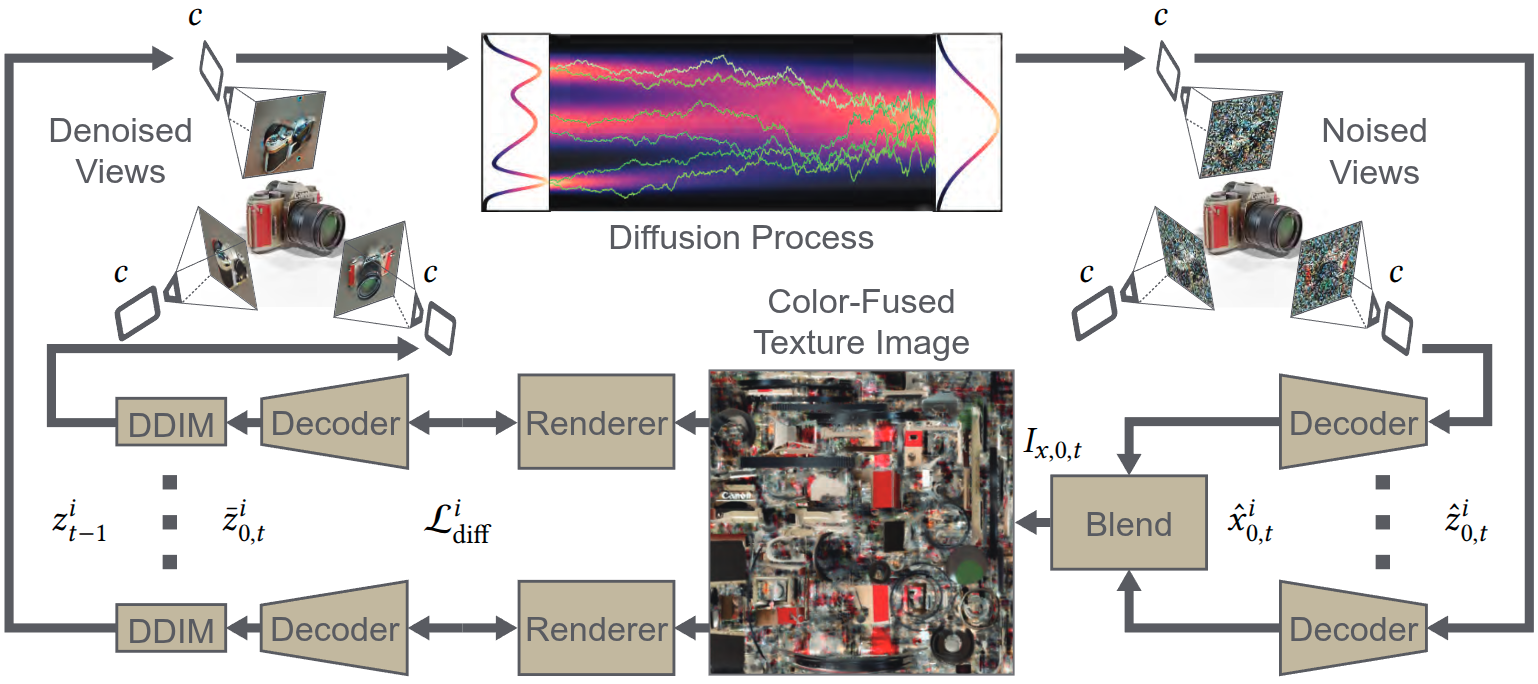}
  \caption{
  An overview of a synchronized multi-view texturing pipeline. Per-view diffusion denoising is coupled through a \emph{Blend} module that fuses per-view textures into a shared texture image, improving cross-view consistency. Figure reproduced from~\cite{zhang2024texpainter}.
  }
  \label{fig:texpainter_arch}
\end{figure}

Another approach is to generate all views simultaneously. Synchronized multi-view texturing methods perform a joint diffusion process across multiple camera views, sampling all views together at each denoising step and aggregating their latent updates onto a common texture representation to promote globally consistent results~\cite{cao2023texfusion,liu2024syncmvd,shao2025mvpainter}. By parallelizing view synthesis, such methods can reduce runtime and improve consistency, since texture evolves jointly across views and overlapping boundary regions are encouraged to agree~\cite{cheng2025mvpaint}. One of the first such approaches, TexFusion~\cite{cao2023texfusion}, introduced multi-view latent fusion in which the diffusion denoiser operates on several rendered views concurrently, and the predicted 2D latent features are aggregated onto a shared latent texture map. The final RGB texture is then recovered by optimizing an intermediate neural color field on decoded renders of that latent texture, yielding globally coherent results without the long per-object optimization used in earlier SDS-based pipelines.

Building on this idea, GenesisTex~\cite{gao2024genesistex} performs diffusion directly in texture space, maintaining one latent texture map per viewpoint during denoising and dynamically aligning them to ensure cross-view coherence. It further introduces style-consistency mechanisms and latent alignment to enforce a unified appearance across viewpoints. A related method, MVPaint~\cite{cheng2025mvpaint}, also generates all views simultaneously using synchronized multi-view diffusion, but follows this with 3D inpainting and UV-space refinement to fill unobserved regions and reduce cross-view inconsistencies, producing seamless high-resolution textures.

Many other recent works adopt synchronized diffusion to improve multi-view consistency. TexPainter~\cite{zhang2024texpainter} enforces multi-view consistent text-to-texture synthesis using a pre-trained latent diffusion model by jointly denoising multiple views. At each DDIM step, the predicted noiseless states are decoded to images, fused in color space into a common texture, and the fusion objective is back-propagated to the view latents, relaxing sequential dependencies and improving cross-view consistency and quality \revone{(Fig.~\ref{fig:texpainter_arch})}. RomanTex~\cite{feng2025romantex} improves geometry- and image-aligned multi-view diffusion for texturing via a 3D-aware rotary positional embedding and a decoupled attention design (reference attention plus multi-view attention), strengthening cross-view coherence before baking to UV space. VCD-Texture~\cite{liu2024vcd} goes further by introducing a 3D--2D collaborative denoising framework that aggregates multi-view 2D latent features into 3D space and rasterizes them back to produce more consistent 2D predictions. This coupling yields highly stable textures, as the 2D diffusion is continuously guided by a holistic 3D representation and vice versa. MaterialMVP~\cite{he2025materialmvp} extends synchronized multi-view diffusion to \emph{PBR} outputs by jointly generating view-consistent albedo and metallic--roughness maps, using consistency-regularized training to suppress illumination artifacts and multi-channel aligned attention to keep the predicted maps spatially aligned. Meanwhile, GenesisTex2~\cite{lu2025genesistex2} extends GenesisTex~\cite{gao2024genesistex} with improved stability and quality by introducing local attention reweighting in the diffusion model’s self-attention layers, ensuring that spatially corresponding patches across views remain strongly correlated. It further merges latent features across views at multiple stages to enforce consistency without compromising diversity.

More recently, SeqTex~\cite{yuan2025seqtex} leverages video diffusion priors to strengthen cross-view coherence in synchronized texturing. It formulates mesh texturing as \emph{sequence generation} and jointly models multi-view renderings and UV textures using geometry-informed attention. By coupling view-space and UV-space generation within a unified framework, it produces complete UV texture maps end-to-end and reduces reliance on post-hoc baking or fusion.

Overall, synchronized approaches demonstrate that parallel multi-view generation greatly enhances texture consistency. By jointly denoising all views through shared latent variables or cross-view interactions, these methods substantially reduce seams. However, UV-space fusion can behave like an averaging operation: if applied too strongly throughout the full denoising trajectory, it can suppress high-frequency details and yield over-smoothed (blurry) textures~\cite{liu2024syncmvd}. As a result, synchronization is often emphasized in early (high-noise) timesteps to align global layout and color, and relaxed in later timesteps to preserve fine details; but once coupling is weakened, per-view updates can drift and residual inconsistencies may still arise, especially in low-overlap or occluded regions, often motivating downstream 3D/UV refinement (\textit{e.g.}, MVPaint~\cite{cheng2025mvpaint}). The main trade-off is higher memory consumption---since multiple views are diffused simultaneously---and, in some cases, the need for custom synchronization mechanisms or additional guidance modules (\textit{e.g.}, GenesisTex~\cite{gao2024genesistex}, FlexiTex~\cite{jiang2025flexitex}). Nevertheless, synchronization leverages the diffusion model’s global coherence to produce seamless textures that are difficult to achieve with strictly view-by-view strategies~\cite{chen2023text2tex,richardson2023texture}.

\paragraph*{Feed-Forward Texturing.} 

\begin{figure}[t]
  \centering
  \includegraphics[width=0.99\linewidth]{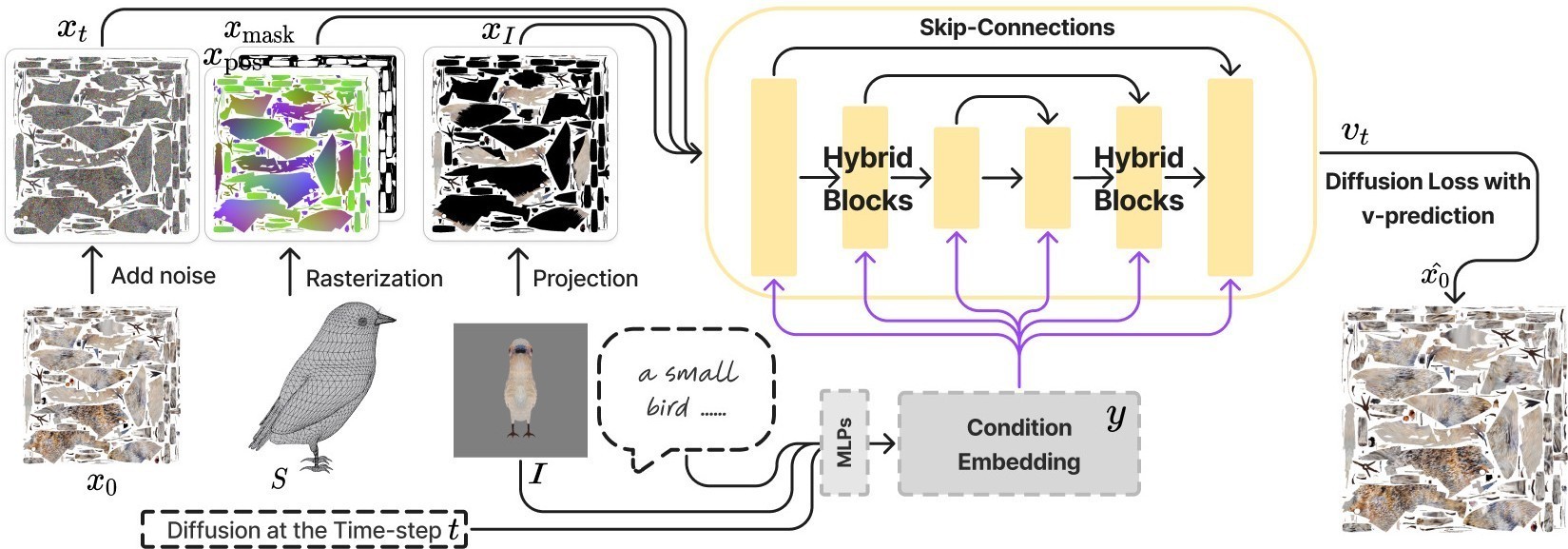}
  \caption{
  A feed-forward texturing pipeline (TEXGen). A diffusion U-Net denoises a latent UV texture atlas conditioned on rasterized mesh cues and text embeddings to generate a UV texture map in a single pass. Figure reproduced from~\cite{yu2024texgen}.
  }
  \label{fig:texgen_arch}
\end{figure}

To avoid per-instance optimization, one solution is to train a diffusion model that directly generates the full texture representation in a single pass, without requiring view-by-view rendering during inference. Many current feed-forward texturing methods operate in UV space, allowing occluded regions to be textured since the full surface is visible to the model. This design avoids view-by-view processing and stitching, but typically requires large datasets of textured 3D objects~\cite{deitke2023objaverse,deitke2023objaversexl}, which have only recently become available (Fig.~\ref{fig:texgen_arch}).

TEXGen~\cite{yu2024texgen} is a 700M-parameter diffusion model trained to generate 1024$\times$1024 UV maps for meshes across diverse categories. Its architecture interleaves 2D convolutions on the UV map with self-attention over a 3D point-cloud representation of the mesh, injecting 3D structural awareness into 2D texture generation. Once trained, TEXGen can synthesize diverse, high-resolution textures in a single forward pass conditioned on text or reference images. Related feed-forward diffusion models also couple 3D and UV-space reasoning: Point-UV Diffusion~\cite{yu2023pointuvdiffusion} adopts a coarse-to-fine design that first diffuses appearance in surface point space to obtain a globally consistent prior, then projects it into UV space and refines a high-resolution texture atlas, helping reduce seams and UV-fragmentation artifacts. More recent feed-forward models continue to broaden this design space. For example, UniTEX~\cite{liang2025unitex} moves beyond purely UV-based processing by lifting texture generation into a unified 3D functional space, while Material Anything~\cite{huang2025material} targets physically based material generation, predicting not only albedo but full PBR material maps for a given 3D object.

Feed-forward models have also explored alternative representations. UV-free Texture Diffusion~\cite{foti2024uv} bypasses UV maps entirely by operating on colored point clouds and using heat diffusion over the mesh surface. Its UV3-TeD framework models each surface as a point set with a learnable latent field, enabling texture generation directly on the mesh without a fixed UV parameterization and thereby avoiding seam- and distortion-related issues associated with UV maps. Likewise, Single-Mesh Diffusion~\cite{mitchel2024fieldlatents} learns field latents attached to mesh vertices and trains a diffusion model to generate these latent codes, which are then decoded to colors. Under a feed-forward generation setting for a given trained asset, this enables rapid, view-consistent synthesis of many texture variants when the geometry is known. Another notable work, AlignTex~\cite{zhang2025aligntex}, addresses the challenge of generating textures that precisely match concept art or reference images of a 3D asset. It employs a diffusion-based two-stage pipeline with alignment losses and transformer-based conditioning to enforce pixel-level correspondence between multi-view artwork and the generated UV map, yielding pixel-precise rather than style-consistent texture synthesis.

Consequently, diffusion-based texturing has rapidly advanced from slow per-mesh optimization~\cite{poole2023dreamfusion,wang2023prolificdreamer} to fast feed-forward generation~\cite{huang2025material}. Iterative methods built on the idea of leveraging 2D diffusion models for 3D texture painting~\cite{chen2023text2tex,richardson2023texture}, and they remain attractive for their simplicity (no model retraining) and controllability (one can intervene at each view). However, they may still produce seams or inconsistent details, since each view is processed sequentially. Synchronized multi-view diffusion~\cite{liu2024syncmvd,cao2023texfusion} significantly alleviates this issue: by treating all views (or the UV map) as a joint diffusion task, these methods encourage the texture to evolve coherently across the surface. The cost is higher memory usage and, in some cases, the need for custom diffusion pipelines, but the benefit is improved quality on challenging cases (intricate objects, full scenes) where view-by-view approaches often struggle with misalignment. Feed-forward approaches~\cite{zhang2025aligntex} push this further by directly mapping text (and other inputs) to the texture domain, often without explicit view-by-view rendering during inference. This yields substantial speedups and opens the door to zero-shot and generative applications, such as texturing large numbers of objects with diversity.

A core challenge for feed-forward models is the distribution of training data. Large diffusion-based texture models are often trained on synthetic datasets or broad 3D asset collections (\textit{e.g.}, ShapeNet~\cite{chang2015shapenet} or Objaverse~\cite{deitke2023objaverse,deitke2023objaversexl}), which may not fully reflect the visual richness of real-world textures. As a result, their outputs may still fall short of the photorealism of real imagery. For instance, a model trained purely on synthetic stylized assets might produce textures with a telltale ``CGI'' look. Current feed-forward models such as Material Anything~\cite{huang2025material} and TEXGen~\cite{huo2024texgen} already exhibit coherent results, but slight domain gaps remain in high-frequency detail and material realism. One promising direction is to combine the strengths of synthetic and real data, \textit{e.g.}, using large synthetic datasets to learn geometry-to-texture alignment while leveraging real image data to improve appearance realism and material properties. Some recent works have moved in this direction by incorporating real-image losses or conditioning signals: Material Anything~\cite{huang2025material} integrates rendering-based losses to encourage physical realism, while FabricDiffusion~\cite{zhang2024fabricdiffusion} uses real fashion photographs as inputs.

\subsection{3D Humans Texturization}
\label{subsec:humans}

Texturing human avatars merits separate discussion, since the underlying assumptions, constraints, and objectives differ from those of general object texturing. First, representation priors differ: many modern human texturing methods build on parametric body models (\textit{e.g.}, SMPL/SMPL-X~\cite{loper2015smpl,pavlakos2019smplx}) with fixed topology and canonical UV atlases, providing consistent surface correspondences across poses~\cite{loper2015smpl,pavlakos2019smplx}. In contrast, general object texturing often lacks such canonical parameterization.

Second, human identity and anatomy impose distinctive constraints: faces, hands, and skin carry fine-grained, identity-defining cues (freckles, lip color, \textit{etc.}) that must be preserved for realism and recognition. Accordingly, many methods incorporate face-specific priors or losses (\textit{e.g.}, a facial UV refinement network or an identity loss) to capture such high-frequency details~\cite{olszewski2017dynamic,deng2018uvgan,wang2019reid}. Moreover, several works segment the texture or mesh into semantic parts and apply part-specific models or losses~\cite{chaudhuri2021semi}.

Third, the data and evaluation protocols are domain-specific: human texturing methods typically train on specialized datasets such as artist-modeled digital humans, 3D scans, or multi-view captures~\cite{liu2024texdreamer,lazova2019texture}, since fully paired ground-truth textures remain scarce. Evaluation also relies on human-centric metrics, including identity preservation (\textit{e.g.}, face-ID or person re-identification~\cite{wang2019reid}) and part-aware image fidelity measures such as PSNR or LPIPS computed on rendered views or aligned texture regions~\cite{chaudhuri2021semi}.

Finally, applications in digital humans demand distinct capabilities: avatars for gaming, virtual try-on, and related interactive settings require textures that are high-fidelity and semantically editable (\textit{e.g.}, ``change the shirt logo to \#10'')~\cite{mir2020clothing,chaudhuri2021semi}. In dynamic settings, even minor misalignments can cause perceptible flicker under motion, breaking realism. Accordingly, human texturing methods often place greater emphasis on semantic controllability and, when animation is involved, temporal consistency of attributes such as clothing patterns~\cite{chaudhuri2021semi,mir2020clothing}.

\paragraph*{Earlier neural pipelines.}
Earlier neural pipelines aim to generate full-body textures from sparse inputs, often a single image, typically leveraging 2D supervision through \emph{UV mapping}. Lazova \etal~\cite{lazova2019texture} reconstruct a full 360$^\circ$ human texture from a single view by fitting an SMPL mesh to the image, projecting visible pixels into its UV atlas, and training a network to inpaint occluded regions. Grigorev~\etal~\cite{grigorev2019coordinate} address pose variation through \emph{coordinate-based inpainting}: they map an input person image into UV space using DensePose~\cite{guler2018densepose} correspondences and train a GAN to fill missing UV regions, helping preserve fine clothing details such as logos and prints under re-posing. A related idea appears in DensePose Transfer~\cite{neverova2018densepose}, which extracts a source UV texture, reprojects it onto a target pose, and refines seams and missing details via a neural network. These UV completion methods---including UV-GAN~\cite{deng2018uvgan} for faces---leverage the maturity of 2D image synthesis in UV space, reducing 3D consistency to a 2D inpainting problem. AUV-Net~\cite{chen2022auvnet} formalizes this further by learning an \emph{aligned} UV parameterization for an entire shape class, mapping semantically corresponding surface points to shared UV coordinates across instances. By training a network to deform each mesh into a common UV template, it achieves dataset-wide texture alignment without manual unwrapping, enabling generative models (GANs or diffusion) trained in this UV space to produce textures transferable across meshes. TexturePose~\cite{pavlakos2019texturepose} does not output an explicit texture map but enforces \emph{texture consistency} during training of a human mesh estimator: if consistent UV textures are predicted across viewpoints, the underlying 3D shape is more likely to be accurate. This is implemented by projecting the image from one view onto the predicted mesh, rendering it to a second view, and comparing against the true image. Although geometry-focused, its core idea---\emph{consistent textures imply consistent geometry}---is conceptually related to later methods that jointly reason about shape and texture.

\paragraph*{Supervised texture generation with human priors.}
Some methods treat human texturing as a direct translation problem from images of people or clothing into a UV texture, often leveraging human parsing or identity cues to handle misalignment. Zhao~\etal~\cite{zhao2020cross} propose a human-parsing-guided texture transfer model in which a single image’s semantic segmentation provides pose and shape cues. During training, they enforce \emph{cross-view consistency} by predicting textures from two views and \emph{exchanging} them to render the opposite view, optimizing a loss between the rendered and input images. This formulation requires no ground-truth 3D textures and enables plausible completion of invisible regions during inference. Wang~\etal~\cite{wang2019reid} address the task using an identity objective, with a person re-identification network serving as a perceptual metric for texture generation. From an input image, they extract a partial texture from visible pixels, render a full-body image, and use a pretrained re-ID CNN to compare this render with the real image. The texture generator is then optimized so that the rendered avatar matches the input identity in feature space. Mir~\etal~\cite{mir2020clothing} address a practical scenario: texturing 3D garments from catalog photos. Given front- and back-view clothing images, their model learns dense correspondences from 2D silhouette coordinates to the garment surface in UV space. Trained on synthetic data with known correspondences, it warps clothing patterns onto a garment template to generate a texture map and supports real-time virtual try-on.

\begin{figure}[t]
  \centering
  \includegraphics[width=0.99\linewidth]{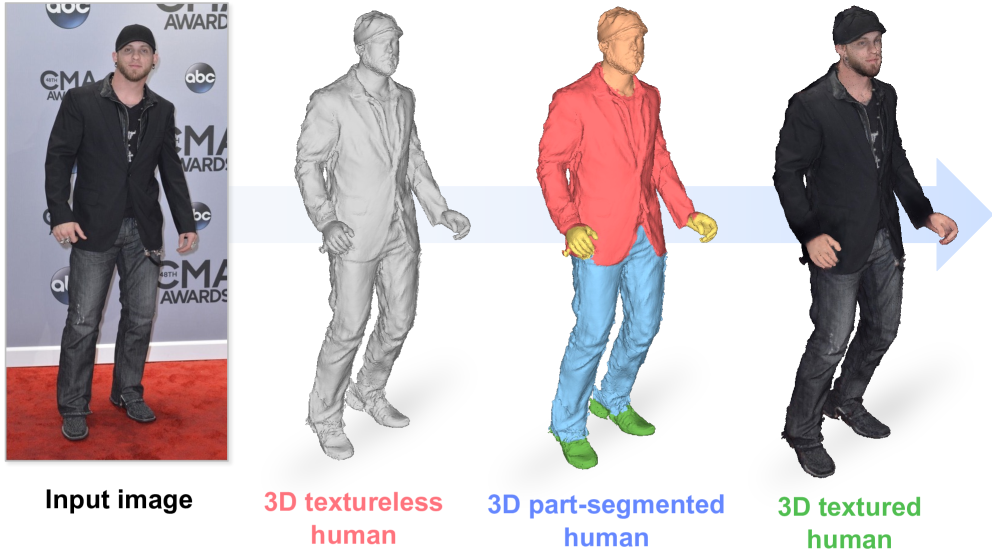}
  \caption{
  Part-aware human texturing. Given a single input image (left), the method segments a textureless human mesh into semantic parts and textures them to produce a detailed 3D human with clean region boundaries (right). Figure adapted from~\cite{nam2025parte}.
  }
  \label{fig:parte}
\end{figure}

Later works extend part-specific texturing. PARTE~\cite{nam2025parte} segments the human mesh into semantic parts (torso, arms, pants, \textit{etc.}) and generates textures for each part to prevent cross-region bleeding. A \emph{PartSegmenter} labels each vertex by part, while a diffusion-based \emph{PartTexturer} fills the corresponding UV regions conditioned on part descriptions, part labels, and neighboring parts. This produces sharper transitions and cleaner textures, improving downstream reconstruction (Fig.~\ref{fig:parte}). Chaudhuri~\etal~\cite{chaudhuri2021semi} propose a semi-supervised framework for generating high-resolution (1024$^2$) editable textures. Their region-adaptive VAE (ReAVAE) learns a latent \emph{style code} for each semantic region (face, shirt, pants, shoes, \textit{etc.}) on the UV map. At generation time, a segmentation mask and optional style vectors guide the synthesis of a consistent texture map, enabling localized edits such as changing only the shirt’s pattern or color. To overcome limited paired data, single-view images are projected into texture space and used as partial supervision. These supervised and weakly supervised methods leverage human-specific cues—such as parsing, keypoints, and identity features—to infer plausible textures from sparse inputs, enabling more flexible generative approaches.

\paragraph*{High-fidelity generation via diffusion-based models.} 

Recent methods move toward more flexible human texture generation using diffusion-based models, with a key focus on maintaining \emph{3D consistency} and \emph{semantic control}. TexDreamer~\cite{liu2024texdreamer} exemplifies this trend as a zero-shot multimodal 3D human texturing system. By adapting a large text-to-image model to a semantic human UV structure using the ATLAS dataset, it generates detailed skin and clothing textures directly in a canonical UV layout, bypassing view rendering. A feature translator maps text or reference-image inputs into the diffusion latent space while conditioning on UV geometry. TexGarment~\cite{liu2025texgarment} targets standalone garments, combining a pre-trained text-to-image diffusion Transformer with 3D structural guidance to generate diverse, 3D-consistent textures. It injects UV position maps and 3D shape cues (\textit{e.g.}, point clouds) to enforce alignment across seams and improve 3D consistency.


Beyond generation, transfer methods also recover reusable materials from real photos. FabricDiffusion~\cite{zhang2024fabricdiffusion} recovers a distortion-free, tileable fabric texture from a single in-the-wild garment photo by ``unwarping'' it into a flat texture map. Trained on synthetic garments with known projected patterns, it infers the underlying tileable material, which can then be mapped to garment UVs and coupled with existing PBR material generation pipelines for high-fidelity rendering and virtual try-on. Make-It-Vivid~\cite{tang2024makeitvivid} explores text-driven texture generation for cartoon characters by adapting a pretrained diffusion model to paired UV maps and text descriptions, and further employs adversarial learning to sharpen details and reduce the synthetic-to-real domain gap.

Finally, methods such as HumanRef~\cite{zhang2024humanref} bridge generative texturing and 3D reconstruction via Ref-SDS, a reference-guided score distillation approach that optimizes a textured 3D human model from a single image. By incorporating image guidance and region-aware attention, HumanRef preserves fine appearance details from the reference image while maintaining cross-view consistency. SiTH~\cite{ho2024sith} takes a related but feed-forward route: it trains an image-conditioned diffusion model to hallucinate unseen back-view appearance from a front view, then feeds the generated views into a mesh reconstruction and texturing pipeline to recover full-body textures. These works highlight how diffusion models and large-scale generative priors support both open-ended texture generation from text or images and strong inpainting of occluded regions where deterministic methods often struggle.

All in all, human texturing has progressed from UV-space completion and transfer~\cite{lazova2019texture,grigorev2019coordinate,mir2020clothing} to powerful diffusion- and transformer-based generation/transfer systems~\cite{liu2024texdreamer,liu2025texgarment,zhang2024fabricdiffusion}, with part-aware designs improving seam handling and editability~\cite{chaudhuri2021semi,nam2025parte}. Yet, several domain-specific hurdles remain. In dynamic settings, temporal consistency under motion remains challenging~\cite{ho2024sith,zhang2024sifu}, and generalization across poses, garments, and body shapes is still limited by scarce paired supervision~\cite{zhao2020cross,pavlakos2019texturepose,wang2019reid}. Facial fidelity and material heterogeneity (skin, hair, fabric) continue to challenge unified modeling~\cite{olszewski2017dynamic,deng2018uvgan,zhang2024fabricdiffusion}, while some diffusion-guided pipelines improve realism at increased computational cost~\cite{liu2024texdreamer,liu2025texgarment}.

Looking forward, we foresee progress toward more transferable canonical UV representations across subjects~\cite{chen2022auvnet,liu2024texdreamer}; part- and seam-aware generators with geometry guidance for complex garments~\cite{nam2025parte,liu2025texgarment}; reference- and text-conditioned editing with identity preservation~\cite{wang2019reid,zhang2024humanref}; and scalable appearance and material modeling that extends beyond RGB~\cite{zhang2024fabricdiffusion}. Achieving these at real-time or near-interactive rates, while maintaining temporal stability, remains a central challenge for neural 3D human texturing.

\subsection{Commercial Systems and Technical Reports}
\label{subsec:industrial_systems}

In addition to the peer-reviewed literature surveyed above, numerous commercial systems, technical reports, and recent preprints have emerged for end-to-end 3D asset generation, often including mesh texturing as a key component~\cite{zhao2025hunyuan3d2,hunyuan3d2025hunyuan3d21,seed3d_report,zhang2024clay,li2025step1x,bensadoun2024meta,rodin_hyper3d_api_docs,tripo_api_docs,meshy_api_docs}. These systems are typically released as hosted products, API services, preprints, or technical write-ups, and therefore differ from academic work in their disclosure level and evaluation practices. We include them here to contextualize research progress with real-world deployment; unless noted otherwise, many sources cited in this subsection are non-peer-reviewed project pages, documentation, or technical reports.

A common theme across these systems is that \emph{texturing is only one component of a larger pipeline} that jointly targets geometry, UVs, and material appearance. For example, Hunyuan3D~\cite{zhao2025hunyuan3d2,hunyuan3d2025hunyuan3d21} describes a two-stage design that first produces a 3D shape and then ``paints'' appearance to obtain a high-fidelity textured asset, with later iterations explicitly incorporating production-oriented material outputs (\textit{e.g.}, PBR-oriented variants) alongside geometry generation. Similarly, Seed3D~\cite{seed3d_report} presents a modular pipeline in which geometry generation is paired with multi-view texture/material synthesis and a subsequent UV completion stage to support downstream use in standard 3D toolchains. CLAY~\cite{zhang2024clay} is a controllable end-to-end 3D asset generator that couples a large 3D geometry model with a multi-view material diffusion module to produce 2K PBR material maps, illustrating how production-oriented systems integrate texturing into the full asset pipeline.


Commercial services such as Rodin~\cite{rodin_hyper3d_api_docs}, Tripo~\cite{tripo_api_docs}, and Meshy~\cite{meshy_api_docs} illustrate how these ideas are operationalized in practice: they emphasize a streamlined user experience (single-image or text-conditioned generation, export-ready textured assets, and direct export to common asset formats) and prioritize robustness across diverse user inputs. While implementation details vary and are not always fully disclosed, these systems broadly align with research trends highlighted throughout this survey: leveraging strong 2D generative priors (diffusion/transformers) for appearance, enforcing multi-view agreement to reduce view-dependent artifacts, and adopting standardized material conventions to ease integration into digital content creation (DCC) tools and real-time renderers.

From a research perspective, these deployments highlight several practical pressures that are likely to shape future texturing work. First, \emph{production constraints}---such as predictable UV conventions, stable material parameterizations, and renderer compatibility---can matter as much as perceptual texture realism. Second, \emph{system-level objectives} such as latency, scalability, failure modes, and controllability become central when texturing is embedded in an end-to-end generation stack rather than studied in isolation. Finally, because many commercial systems are distributed as services, their training data curation, preprocessing, and evaluation protocols are often less transparent than in academic releases, motivating reproducible benchmarks and standardized reporting for fair comparison.

\begin{table*}[p]  


\caption{
A summary of representative works in \textbf{Neural 3D Mesh Texturing}.
Each work is characterized by \textbf{Model Type} (the neural model or backbone used),
\textbf{Guidance} (the Stylistic Guidance controlling texture appearance),
\textbf{Training Strategy} (whether the model is Pre-trained, Fine-tuned, or Custom trained for the texturing task),
\textbf{Generation Strategy} (how textures are generated: Optimization, Iterative, Synchronized, or Feed-forward),
and output \textbf{Texture Type} (RGB textures, often with baked-in lighting effects, or disentangled PBR materials).
}

\centering
\resizebox{0.99\textwidth}{!}{
\begin{tabular}{lccccc} 
\toprule
\textbf{Methods}  & \textbf{Model Type} & \textbf{Guidance}  & \textbf{Training Strategy} & \textbf{Generation Strategy} & \textbf{Texture Type} \\
\midrule

\rowcolor[HTML]{E3F2FD}
TextureFields~\cite{oechsle2019texturefields} & Neural fields & Uncond./Image & Custom & Feed-forward & RGB textures  \\
\rowcolor[HTML]{FBFEFF}
NMR~\cite{kato2018nmr} & VGG & Image & Pre-trained & Optimization & RGB textures  \\

\arrayrulecolor{gray}\cmidrule(lr){1-6}
\arrayrulecolor{black}

\rowcolor[HTML]{E8F5E9}
Huang \etal~\cite{huang2020adversarial} & GAN & Image & Custom & Optimization & RGB textures  \\
\rowcolor[HTML]{FBFEFC}
LTG~\cite{yu2021learning_texture_generators} & GAN & Unconditional & Custom & Feed-forward & RGB textures  \\
\rowcolor[HTML]{E8F5E9}
Texturify~\cite{siddiqui2022texturify} & GAN & Unconditional & Custom & Feed-forward & RGB textures  \\
\rowcolor[HTML]{FBFEFC}
Mesh2Tex~\cite{bokhovkin2023mesh2tex} & GAN+Neu. fields & Uncond./Image & Custom & Feed-forward+Opt. & RGB textures  \\

\rowcolor[HTML]{E8F5E9}
ShaDDR~\cite{chen2023shaddr} & GAN & 3D shape & Custom & Feed-forward & RGB textures  \\
\rowcolor[HTML]{FBFEFC}
3DStyleNet~\cite{yin20213dstylenet} & GAN & 3D shape & Pre-trained+Custom & Optimization & RGB textures  \\
\rowcolor[HTML]{E8F5E9}
SPSG~\cite{dai2021spsg} & GAN & Image & Custom & Feed-forward & RGB textures  \\

\rowcolor[HTML]{FBFEFC}
AUV-Net~\cite{chen2022auvnet} & GAN & Uncond./Image/3D shape & Custom & Feed-forward+Opt. & RGB textures  \\
\rowcolor[HTML]{E8F5E9}
Chaudhuri \etal~\cite{chaudhuri2021semi} & VAE+GAN & Uncond./Image & Pre-trained+Custom & Feed-forward & RGB textures  \\
\rowcolor[HTML]{FBFEFC}
Lazova \etal~\cite{lazova2019texture} & GAN & Image & Custom & Feed-forward & RGB textures  \\
\rowcolor[HTML]{E8F5E9}
Grigorev \etal~\cite{grigorev2019coordinate} & GAN & Image & Custom & Feed-forward & RGB textures  \\
\rowcolor[HTML]{FBFEFC}
Neverova \etal~\cite{neverova2018densepose} & GAN & Image & Custom & Feed-forward & RGB textures  \\
\rowcolor[HTML]{E8F5E9}
UV-GAN~\cite{deng2018uvgan} & GAN & Image & Custom & Feed-forward & RGB textures  \\

\arrayrulecolor{gray}\cmidrule(lr){1-6}
\arrayrulecolor{black}

\rowcolor[HTML]{FFFEFC}
Dream Fields~\cite{jain2022dreamfields} & CLIP & Text & Pre-trained+Custom & Optimization & RGB textures \\
\rowcolor[HTML]{FFF3E0}
Text2Mesh~\cite{michel2022text2mesh} & CLIP & Text & Pre-trained+Custom & Optimization  & RGB textures  \\
\rowcolor[HTML]{FFFEFC}
X-Mesh~\cite{ma2023xmesh} & CLIP & Text & Pre-trained+Custom & Optimization & RGB textures  \\
\rowcolor[HTML]{FFF3E0}
TANGO~\cite{chen2022tango} & CLIP & Text & Pre-trained+Custom & Optimization & PBR materials  \\
\rowcolor[HTML]{FFFEFC}
CLIP-Mesh~\cite{khalid2022clipmesh} & CLIP & Text & Pre-trained & Optimization & RGB textures  \\

\arrayrulecolor{gray}\cmidrule(lr){1-6}
\arrayrulecolor{black}

\rowcolor[HTML]{F3E5F5}
TEXTure~\cite{richardson2023texture} & Diffusion (2D) & Text & Pre-trained & Iterative & RGB textures \\
\rowcolor[HTML]{FEFAFF}
Text2Tex~\cite{chen2023text2tex} & Diffusion (2D) & Text & Pre-trained & Iterative & RGB textures \\
\rowcolor[HTML]{F3E5F5}
TexFusion~\cite{cao2023texfusion} & Diffusion (2D)+Neu. fields & Text & Pre-trained+Custom & Synchronized + Opt. & RGB textures \\
\rowcolor[HTML]{FEFAFF}
Paint3D~\cite{zeng2024paint3d} & Diffusion (2D+UV) & Text/Image & Pre-trained+Custom & Iterative (+UV refine) & RGB textures \\  
\rowcolor[HTML]{F3E5F5}
SyncMVD~\cite{liu2024syncmvd} & Diffusion (2D) & Text & Pre-trained & Synchronized & RGB textures \\
\rowcolor[HTML]{FEFAFF}
MVPaint~\cite{cheng2025mvpaint} & Diffusion (2D+UV) & Text & Pre-trained+Custom & Synchronized (+refine) & RGB textures \\
\rowcolor[HTML]{F3E5F5}
TexPainter~\cite{zhang2024texpainter} & Diffusion (2D) & Text & Pre-trained & Synchronized+Opt. & RGB textures \\
\rowcolor[HTML]{FEFAFF}
VCD-Texture~\cite{liu2024vcd} & Diffusion (2D) & Text & Pre-trained & Synchronized (+refine) & RGB textures \\
\rowcolor[HTML]{F3E5F5}
TexGen~\cite{huo2024texgen} & Diffusion (2D) & Text & Pre-trained & Synchronized & RGB textures \\
\rowcolor[HTML]{FEFAFF}
GenesisTex~\cite{gao2024genesistex} & Diffusion (2D) & Text & Pre-trained & Synchronized (+refine) & RGB textures \\
\rowcolor[HTML]{F3E5F5}
GenesisTex2~\cite{lu2025genesistex2} & Diffusion (2D) & Text & Pre-trained & Synchronized & RGB textures \\
\rowcolor[HTML]{FEFAFF}
FlexiTex~\cite{jiang2025flexitex} & Diffusion (2D) & Text/Image & Pre-trained & Synchronized & RGB textures \\
\rowcolor[HTML]{F3E5F5}
RoomPainter~\cite{huang2025roompainter} & Diffusion (2D) & Text & Pre-trained & Synchronized & RGB textures \\
\rowcolor[HTML]{FEFAFF}
InstanceTex~\cite{yang2024instancetex} & Diffusion (2D)+Neu. fields & Text & Fine-tuned+Custom & Iterative+Opt. & RGB textures \\
\rowcolor[HTML]{F3E5F5}
RoomTex~\cite{wang2024roomtex} & Diffusion (2D) & Text & Pre-trained & Iterative & RGB textures \\
\rowcolor[HTML]{FEFAFF}
Make-A-Texture~\cite{xiang2025make} & Diffusion (2D) & Text & Pre-trained & Iterative & RGB textures \\
\rowcolor[HTML]{F3E5F5}
AlignTex~\cite{zhang2025aligntex} & Diffusion (2D) & Image & Fine-tuned & Feed-forward+Sync. & RGB textures \\
\rowcolor[HTML]{FEFAFF}
TEXGen (700M)~\cite{yu2024texgen} & Diffusion (UV) & Text/Image & Custom & Feed-forward & RGB textures \\
\rowcolor[HTML]{F3E5F5}
Single-Mesh DM~\cite{mitchel2024fieldlatents} & Diffusion (on surface) & Unconditional & Custom & Feed-forward & RGB textures \\
\rowcolor[HTML]{FEFAFF}
Diffu. Tex. Paint.~\cite{dtp} & Diffusion (2D) & Image/Brush & Fine-tuned & Feed-forward (interactive) & RGB textures \\
\rowcolor[HTML]{F3E5F5}
TextureDreamer~\cite{yeh2024texturedreamer} & Diffusion (2D) & Image & Fine-tuned+Custom & Optimization & PBR materials \\
\rowcolor[HTML]{FEFAFF}
StyleTex~\cite{xie2024styletex} & Diffusion (2D) & Image & Pre-trained & Optimization & RGB textures \\
\rowcolor[HTML]{F3E5F5}
FlashTex~\cite{deng2024flashtex} & Diffusion (2D) & Text & Fine-tuned+Custom & Synchronized+Opt. & RGB textures \\
\rowcolor[HTML]{FEFAFF}
DreamMat~\cite{zhang2024dreammat} & Diffusion (2D) & Text & Fine-tuned+Custom & Optimization & PBR materials \\
\rowcolor[HTML]{F3E5F5}
MaPa~\cite{zhang2024mapa} & Diffusion (2D) & Text & Fine-tuned+Custom & Optimization & PBR materials \\
\rowcolor[HTML]{FEFAFF}
Decorate3D~\cite{guo2023decorate3d} & Diffusion (2D) & Text/Image & Pre-trained+Custom & Optimization & RGB textures \\
\rowcolor[HTML]{F3E5F5}
3D Paintbrush~\cite{decatur2024paintbrush} & Diffusion (2D) & Text & Pre-trained+Custom & Optimization & RGB textures \\
\rowcolor[HTML]{FEFAFF}
EASI-Tex~\cite{perla2024easitex} & Diffusion (2D) & Image & Pre-trained & Iterative & RGB textures \\
\rowcolor[HTML]{F3E5F5}
Paint-it~\cite{kim2024paintit} & Diffusion (2D) & Text & Pre-trained+Custom & Optimization & PBR materials \\
\rowcolor[HTML]{FEFAFF}
Fantasia3D~\cite{chen2023fantasia3d} & Diffusion (2D) & Text & Pre-trained+Custom & Optimization & PBR materials \\
\rowcolor[HTML]{F3E5F5}
Material Anything~\cite{huang2025material} & Diffusion (2D+UV) & Text/3D Mesh & Fine-tuned+Custom & Feed-forward & PBR materials \\
\rowcolor[HTML]{FEFAFF}
TexDreamer~\cite{liu2024texdreamer} & Diffusion (2D/UV) & Text/Image & Fine-tuned & Feed-forward & RGB textures \\  
\rowcolor[HTML]{F3E5F5}
Make-It-Vivid~\cite{tang2024makeitvivid} & Diffusion (2D/UV) & Text & Fine-tuned & Feed-forward & RGB textures \\  
\rowcolor[HTML]{FEFAFF}
FabricDiffusion~\cite{zhang2024fabricdiffusion} & Diffusion (2D) & Image & Fine-tuned & Feed-forward (UV/tile) & RGB textures \\  
\rowcolor[HTML]{F3E5F5}
TexGarment~\cite{liu2025texgarment} & Diffusion (2D/UV) & Text & Fine-tuned+Custom & Feed-forward & RGB textures \\  
\rowcolor[HTML]{FEFAFF}
Point-UV Diffusion~\cite{yu2023pointuvdiffusion} & Diffusion (Point+UV) & Uncond./Text/Image & Custom & Feed-forward & RGB textures \\
\rowcolor[HTML]{F3E5F5}
SeqTex~\cite{yuan2025seqtex} & Diffusion (Video+UV) & Text/Image & Fine-tuned & Synchronized & RGB textures \\
\rowcolor[HTML]{FEFAFF}
RomanTex~\cite{feng2025romantex} & Diffusion (2D) & Image & Fine-tuned & Synchronized & RGB textures \\
\rowcolor[HTML]{F3E5F5}
MaterialMVP~\cite{he2025materialmvp} & Diffusion (2D) & Image & Fine-tuned & Synchronized & PBR materials \\
\rowcolor[HTML]{FEFAFF}
CLAY~\cite{zhang2024clay} & Diffusion (2D) & Text/Image & Custom & Synchronized & PBR materials \\
\rowcolor[HTML]{F3E5F5}
VideoMat~\cite{munkberg2025videomat} & Diffusion (Video+PBR) & Text/Image & Fine-tuned & Synchronized & PBR materials \\
\rowcolor[HTML]{FEFAFF}
DreamPBR~\cite{xin2025dreampbr} & Diffusion (2D+PBR) & Text+Multimodal & Fine-tuned+Custom & Feed-forward & PBR materials \\

\bottomrule
\end{tabular}
} 
\label{tab:all}
\end{table*}

\section{Datasets and Evaluation Metrics}
\label{sec:datasets-metrics}

This section reviews the datasets and evaluation metrics commonly used in neural methods that \emph{directly} texture 3D meshes. We focus on datasets that offer textured meshes or enable reliable supervision for mesh texturing, as well as standard metrics for evaluating the fidelity, consistency, and usability of textured assets.

\subsection{Datasets}
\label{subsec:datasets}

High-quality textured 3D meshes remain relatively scarce compared to geometry-only repositories. As a result, most works rely on one or more of three data sources: (i) curated datasets of textured meshes; (ii) geometry-only collections combined with external image datasets or procedural materials; and (iii) large-scale, web-scraped 3D asset libraries with varying texture quality and licensing. Below, we summarize representative datasets grouped by content type.

\subsubsection{Mesh Datasets}
\paragraph*{General Object Mesh Datasets.} 

Early repositories such as ModelNet~\cite{wu20153d} and ShapeNet~\cite{chang2015shapenet} contain thousands of 3D CAD models but are mostly untextured or only sparsely textured. Similarly, Thingi10K~\cite{zhou2016thingi10k} includes 10,000 3D-printable models, emphasizing geometric diversity over material realism. The Princeton COSEG dataset~\cite{wang2012cosegdataset} provides segmented 3D models across object categories for shape analysis, but with little or no texture information. PASCAL‐3D+~\cite{xiang_wacv14} bridges 2D and 3D domains by aligning 3D CAD models with real images for object detection and pose estimation, but it does not provide high-quality textured meshes. In contrast, PhotoShape~\cite{park2018photoshape} augmented a subset of ShapeNet~\cite{chang2015shapenet} with photorealistic materials inferred from internet photos, yielding a valuable resource for learning-based rendering and texturing research.

\begin{figure}[t]
  \centering
  \includegraphics[width=0.99\linewidth]{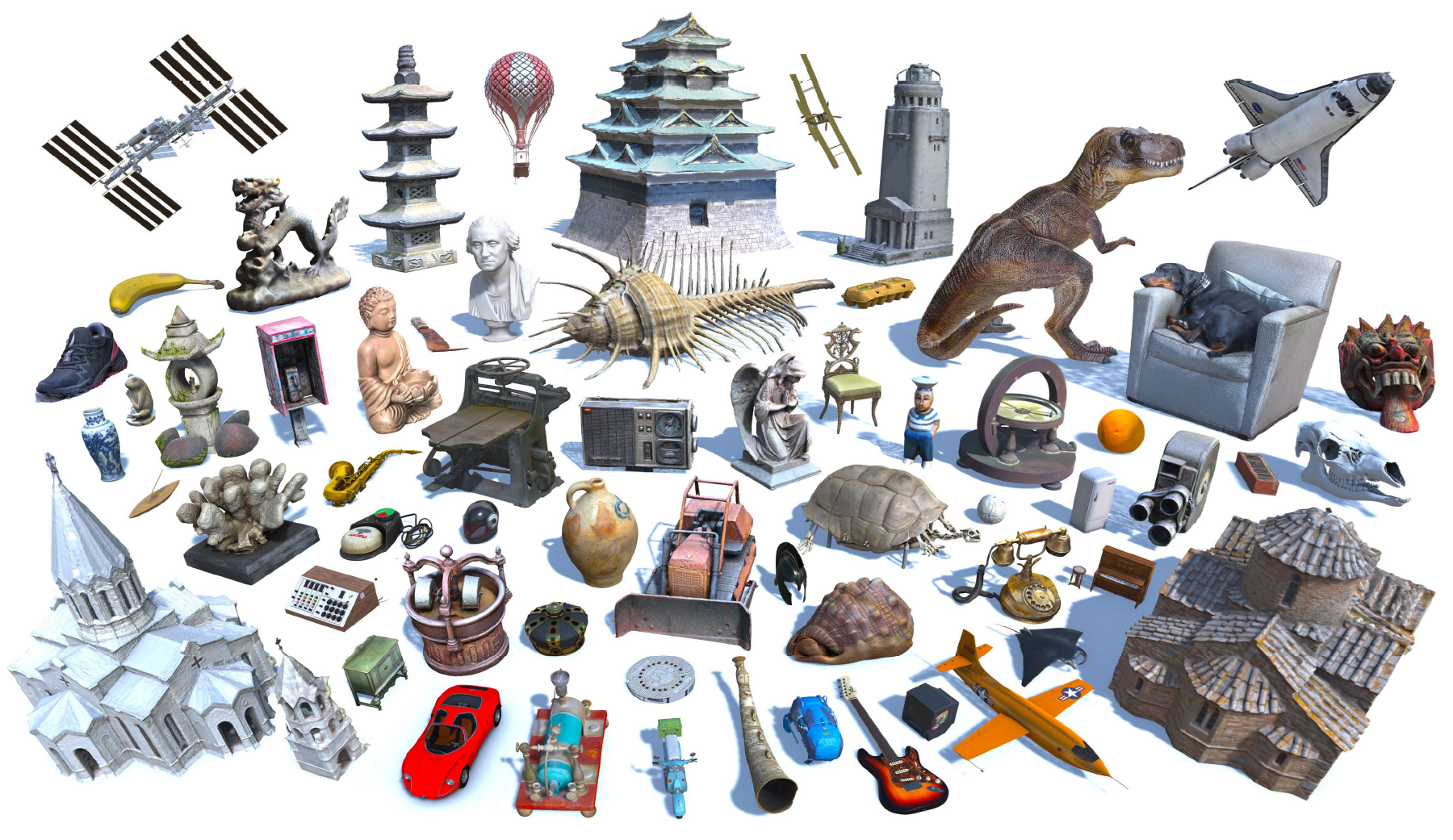}
  \caption{
  Objaverse-XL provides over 10M 3D objects spanning diverse categories, enabling large-scale training and benchmarking for mesh texturing. Figure reproduced from~\cite{deitke2023objaversexl}.
  }
  \label{fig:objaverseXL}
\end{figure}

Recent datasets explicitly emphasize diverse, high-quality object textures. 3D-FUTURE~\cite{fu20213dfuture} contains about 10,000 furniture CAD models with high-resolution textures and rich annotations, targeting household objects in indoor scenes. It is often paired with the 3D-FRONT scene dataset~\cite{fu20213dfront}, which provides complete room layouts furnished with 3D-FUTURE assets to support indoor-scene synthesis and related texture-transfer research. The Amazon-Berkeley Objects (ABO) dataset~\cite{collins2022abo} likewise offers thousands of product models with realistic geometry and physically based materials, helping bridge synthetic and real-world object understanding. At larger scale, Objaverse~\cite{deitke2023objaverse} and its successor Objaverse-XL~\cite{deitke2023objaversexl} aggregate millions of 3D models across highly diverse categories; although not all models feature high-quality textures \revone{(Fig.~\ref{fig:objaverseXL})}, these web-scale collections have become valuable resources for pre-training and large-scale generative 3D/texturing research. The recently introduced TexVerse dataset~\cite{zhang2025texverseuniverse3dobjects} further advances both scale and quality, curating over 850K unique 3D objects with high-resolution textures, including over 158K with full PBR maps, as well as specialized subsets of rigged and animated textured models, making it a promising resource for learning neural texturing across object types.

For research requiring scenes, the Matterport3D dataset~\cite{Matterport3D} provides 90 real indoor scenes with RGB-D captures, surface reconstructions, and textured meshes, making it a valuable resource for methods that leverage scene-level geometry and appearance. Some synthetic datasets focus on specialized use cases; for example, Houses3K~\cite{peralta2020next} contains 3,000 procedurally generated house models with multiple texture variants, originally created to train next-best-view policies for 3D reconstruction. In summary, a broad spectrum of mesh datasets now exists, from early geometry-only repositories to modern collections with curated textures, which can be combined to train and evaluate neural texturing methods.

\paragraph*{3D Humans and Garments.} 

Textured 3D human models present unique challenges, and relatively few large public datasets are available. Several commercial or restricted 3D human resources exist, including RenderPeople~\cite{renderpeople_2025}, Triplegangers~\cite{triplegangers_2025}, Treedy~\cite{treedys_2025}, and Twindom~\cite{twindom_2025}. These resources provide high-fidelity scanned humans or avatar assets with realistic textures, but they are typically proprietary or only partially accessible. Recent academic efforts have instead focused on larger-scale human texture datasets. For example, TexDreamer’s ATLAS dataset~\cite{liu2024texdreamer}, described as the ``largest high-resolution 3D human texture dataset'', offers diverse human UV textures for zero-shot generative modeling. Although ATLAS has not been fully released, it enabled TexDreamer’s high-quality results in 3D human texturing. For clothed virtual characters, the 3DBiCar dataset~\cite{luo2023rabit} provides 1,500 fully textured cartoon-style human meshes spanning 15 character species.

Many works also leverage 2D data to compensate for the scarcity of textured 3D human scans. For instance, single- or multi-view images of people, as well as curated internet photos of clothing, are often used to texture 3D human models~\cite{mir2020clothing,cha2023singleimagetex}. Therefore, while large-scale public datasets of textured 3D humans remain limited, the community is gradually assembling both realistic (scanned) and synthetic (artistic or generated) texture collections to advance learning-based texturing of people and garments.

\subsubsection{Image Datasets}

In addition to 3D datasets, many methods rely on 2D image datasets for learning or evaluation, especially when only geometry is available. For example, image-guided texture generation for specific object categories uses object image collections such as BrnoCompSpeed~\cite{sochor2019brnocompspeed}, which includes over 20k real car images with ground-truth annotations, and CompCars~\cite{yang2015compcars}, which provides 214k car images spanning make and model variations, to synthesize realistic car paint or decal textures. Bird texture transfer methods likewise employ the CUB-200-2011 dataset~\cite{wah2011caltech} to map fine-grained feather patterns onto 3D bird models. Generic texture descriptors are often learned from the Describable Textures Dataset (DTD)~\cite{cimpoi2014describing}, containing 5,640 images of patterned textures labeled with human-descriptive attributes. For texture super-resolution or recovering high-frequency details, high-resolution datasets such as DIV2K~\cite{agustsson2017ntire}, with 1,000 2K-resolution images, are commonly used as benchmarks~\cite{richard2019multi}.

Several large-scale human-centric image datasets have been instrumental in learning to texture people and clothing. The DeepFashion dataset~\cite{liucvpr16DeepFashion}, with over 800k clothing images annotated with attributes and landmarks, and Market-1501~\cite{Zheng_2015_ICCV}, comprising 32,000+ multi-camera person images, serve as abundant 2D sources of in-the-wild clothing appearance and are often used as supervision or priors for methods that transfer garment appearance onto 3D human meshes. The MVC dataset~\cite{liu2016mvc} provides multi-view photographs of about 37,000 clothing items with attribute labels, enabling learning of view-invariant clothing texture representations. For faces, the FFHQ dataset~\cite{karras2019stylegan} (70,000 high-quality human face images) is often used to learn realistic facial appearance priors, including for facial texture or albedo synthesis. Two additional datasets link images to human UV texture maps: DensePose-COCO~\cite{guler2018densepose} augments COCO images~\cite{lin2014microsoft} with dense body-surface correspondences and part-specific UV coordinates, facilitating supervised texture transfer and completion, while the CMU Multi-PIE dataset~\cite{gross2008multipie} provides multiview facial images under varying illumination, supporting the learning of consistent UV facial textures across viewpoints.

Finally, beyond established datasets, many works curate task-specific data from internet imagery to suit particular applications---\textit{e.g.}, assembling custom clothing-image collections for texture exemplars or constructing custom 2D--3D correspondences~\cite{zhang2024fabricdiffusion,yu2021learning_texture_generators}. Such one-off datasets highlight the community’s continuing need to fill gaps in data availability.

\subsubsection{Dataset Preprocessing and Filtering}

\paragraph*{Meshes.}
Large-scale mesh repositories collected from the web (\textit{e.g.}, Objaverse/Objaverse-XL~\cite{deitke2023objaverse,deitke2023objaversexl}) exhibit substantial variation in geometry, UV parameterizations, texture conventions, and asset completeness, making careful preprocessing and filtering important for stable training and fair evaluation. For instance, even widely used collections contain many assets with missing or low-resolution textures, and reported cleaning pipelines often enforce minimum texture-resolution thresholds, exclude assets with restrictive usage tags (\textit{e.g.}, terms related to ``NoAI''), and restrict the dataset to redistributable Creative Commons licenses before standardizing formats such as \texttt{.glb} for downstream use~\cite{zhang2025texverseuniverse3dobjects,li2025step1x,shao2025mvpainter}. Beyond basic sanity checks, practical mesh filtering commonly includes: (i) \emph{geometry quality} checks (degenerate faces, corrupted topology, extreme scale/units, missing normals/material assignments); and (ii) \emph{UV validity} checks (presence of UVs, invalid/NaN coordinates, severe overlaps, excessive fragmentation, or extremely low texel density), since UV pathologies can directly translate into artifacts or unstable supervision when learning in UV space.

Recent texturing pipelines provide concrete examples of such curation. TexGen~\cite{huo2024texgen} reports that web meshes come with inconsistent ``texture structures'' and quality; it filters poor texture cases, re-unwraps meshes to a new parameterization, and bakes diffuse color into the new UVs to obtain a consistent training representation. For PBR-oriented supervision, Material Anything~\cite{huang2025material} constructs Material3D by filtering Objaverse~\cite{deitke2023objaverse} meshes to retain only assets with a sufficiently complete material-map set (\textit{e.g.}, base color, roughness, metallic, and bump), then re-unwrapping and consolidating parts to produce UV-ready training data. Such steps also reduce confounds in evaluation: improvements in a learned model should not be attributable to inconsistent UV layouts, missing maps, or broken assets.

A key additional consideration is \emph{PBR availability and imbalance}. High-quality multi-map materials are substantially less common than albedo-only textures or shaded renders; for example, TexVerse~\cite{zhang2025texverseuniverse3dobjects} curates 858K high-resolution textured models, but only a subset (158K) contains PBR materials under standard metalness--roughness or specular--glossiness workflows with the requisite roughness/glossiness and metalness/specular channels. This scarcity creates an inherent data imbalance between RGB-only supervision and fully relightable PBR supervision, and motivates reporting dataset statistics (\#assets with UVs, \#with albedo-only, \#with full PBR sets, map resolutions) as well as clearly stating conversion choices (\textit{e.g.}, metalness--roughness vs.\ specular--glossiness, map packing, and color-space conventions). Overall, documenting preprocessing decisions helps readers interpret results and improves reproducibility across datasets and renderers.

\paragraph*{Images.}
When methods rely on reference or conditioning images (\textit{e.g.}, as references~\cite{perla2024easitex}), basic curation helps prevent the image domain from becoming a hidden source of artifacts and bias. Common steps include (i) foreground isolation via segmentation/matting (to avoid background leakage into the synthesized texture), (ii) cropping to a region of interest and resizing to a consistent resolution, and (iii) filtering for excessive occlusion, extreme viewpoints, or low sharpness~\cite{perla2024easitex,siddiqui2022texturify}. Some pipelines additionally leverage region-level crops or masks to isolate the relevant appearance signal (\textit{e.g.}, material/print patterns) and reduce interference from background clutter or occlusions; for example, FabricDiffusion~\cite{zhang2024fabricdiffusion} conditions on a clothing image together with region captures of its fabric materials and prints to extract normalized textures/prints, which are then mapped onto the target garment UVs. Overall, while image preprocessing is typically lighter-weight than mesh filtering, these steps improve robustness and reduce spurious texture transfer driven by backgrounds or irrelevant context.

\subsection{Evaluation Metrics}
\label{subsec:eval-metrics}

For neural 3D mesh texturing, ground-truth data are often unavailable. As a result, acceptable outputs vary by goal (\textit{e.g.}, transfer, alignment, or conditioned synthesis), leading to task-specific and non-standardized evaluations. In practice, most works report a combination of image-based and asset-level criteria, assessing \emph{distribution-level realism}, \emph{per-instance fidelity}, and \emph{semantic alignment}, often complemented by user studies. For controlled comparison, evaluations are typically conducted on multi-view renderings with fixed camera sets and lighting or environment maps, with background masking when comparing to real photographs to isolate object appearance. Robustness is also tested across mesh types (organic \textit{vs.}\ CAD-like, generated \textit{vs.}\ artist-authored) and is often accompanied by runtime analysis for practical relevance~\cite{cao2023texfusion,richardson2023texture,xie2024styletex,liu2024syncmvd}. Below, we mark $\uparrow$ where higher metric values indicate better performance and $\downarrow$ where lower values are better.

\paragraph*{Appearance Alignment.}
To measure how closely the distribution of generated textured images matches real images or ground-truth textures, authors often rely on GAN-style distribution metrics. The Fréchet Inception Distance (FID; $\downarrow$)~\cite{heusel2017fid} computes the 2-Wasserstein distance~\cite{villani2009ot,dowson1982frechet,gelbrich1990w2} between Gaussian approximations of deep feature distributions (typically extracted using an Inception-V3 network~\cite{szegedy2016rethinking}) for rendered results and reference images. Lower FID indicates that the generated distribution is closer to the reference distribution. Similarly, the Kernel Inception Distance (KID; $\downarrow$)~\cite{binkowski2018kid} is an unbiased MMD-based~\cite{gretton2012kernel} metric comparing sets of Inception features; it is often preferred for smaller sample sizes. The Inception Score (IS; $\uparrow$)~\cite{salimans2016gantrain} measures both confidence and diversity by feeding generated images into a pretrained classifier: sharp, class-consistent predictions for individual images together with high entropy over the marginal label distribution yield a higher IS. These distribution-level metrics are most common when a unique ground-truth texture is unavailable (\textit{e.g.}, style transfer) and realism or diversity should match a target distribution. We note, however, that their usefulness can vary: \textit{e.g.}, FID and IS are less sensitive to spatial alignment or fine geometric detail, and reliable estimates generally require sufficiently large sample sizes to reduce statistical noise. Therefore, they are usually supplemented by more granular metrics as described next.

\paragraph*{Instance-Level Reconstruction Fidelity.}
When a specific target texture or image is available for each 3D model (\textit{e.g.} in texture super-resolution or image-based texturing tasks), evaluation can treat the problem as an image reconstruction task, and standard pixel-wise or structural metrics from image processing can be employed. Peak Signal-to-Noise Ratio (PSNR; $\uparrow$)~\cite{zhang2018lpips} and the Structural Similarity Index (SSIM; $\uparrow$)~\cite{wang2004ssim,wang2003multiscale} are frequently reported to quantify low-level fidelity---higher PSNR or SSIM on rendered views indicates that the synthesized texture preserves more detail and structure from the ground truth. Because these pointwise metrics often fail to reflect perceptual quality, many works also use learned perceptual distances; for example, the Learned Perceptual Image Patch Similarity (LPIPS; $\downarrow$)~\cite{zhang2018lpips}, which measures distance using deep features, is a popular gauge of human-perceived closeness (lower LPIPS means more perceptually alike), especially in texture super-resolution. If the task involves semantic or part-aware accuracy, and ground-truth regions are available (such as in Texture Alignment~\cite{chen2022auvnet}), Intersection-over-Union (IoU; $\uparrow$) or other overlap measures can be used to evaluate how well the predicted textures align with those regions.

\paragraph*{Prior-Guided Semantic Alignment.} 
In many neural texturing scenarios, the goal is for the output texture to match some input descriptions such as a text prompt, a style image, or the visual characteristics of a source domain. Here, evaluation leverages pretrained models as ``priors'' to score alignment. A common choice is CLIP~\cite{radford2021clip}: the CLIP Score (image--text cosine similarity; $\uparrow$)~\cite{hessel2021clipscore} measures how well a rendered image of the textured mesh matches a given text prompt in CLIP’s joint embedding space, while CLIP-Var (image--image cosine similarity; $\uparrow$)~\cite{li2024learning} evaluates the consistency of multi-view renderings of the textured mesh. Likewise, for text-to-texture tasks, authors also report CLIP R-Precision ($\uparrow$)~\cite{park2021compt2i}, which measures whether the correct text prompt is retrieved among a set of distractors using CLIP embeddings. In practice, this retrieval-based metric serves as a reasonable proxy for human judgments of text--image correspondence~\cite{liu2024vcd}. Additionally, some works report the \emph{CLIP--Aesthetic Score} ($\uparrow$), which is a no-reference regressor over CLIP embeddings trained to predict perceived visual appeal. This complements semantic alignment metrics but does not directly assess alignment with a prompt~\cite{schuhmann2022aestheticpredictor}.

Beyond CLIP, other perceptual or semantic priors also inform evaluation. Some works compute feature-space distances such as Feature-$\ell_1/\ell_2$ ($\downarrow$) between rendered outputs and target images using pretrained CNN features (\textit{e.g.}, VGG~\cite{simonyan2015vgg}). These distances provide a perceptual or semantic fidelity measure and are widely used in settings such as style transfer and image-based optimization~\cite{gatys2016styletransfer,kato2018nmr}. No-reference image quality models are also repurposed to assess textures: the Blind/Referenceless Image Spatial Quality Evaluator (BRISQUE; $\downarrow$)~\cite{mittal2012brisque} measures deviations from natural scene statistics, so lower BRISQUE generally indicates more natural-looking outputs. Similarly, Generative Image Quality Assessment (GIQA; $\uparrow$)~\cite{gu2020giqa} uses a learned model to predict the quality of each generated image. A higher GIQA score indicates better per-image quality, complementing FID’s dataset-level view.

Very recently, researchers have begun leveraging multimodal large language models for evaluation. One example is the MLLM Score ($\uparrow$)~\cite{huang2023t2icompbench,huang2025t2icompbench++}, which prompts a vision--language model to assess how well an image satisfies a textual instruction or description. Finally, we note that certain metrics target specific texture properties. For instance, TexTile ($\downarrow$)~\cite{rodriguezpardo2024textile} is a learned metric designed to quantify the tileability of a texture (\textit{i.e.}, whether it can repeat without visible seams)~\cite{zhang2024fabricdiffusion}. In summary, prior-guided metrics enable evaluation of semantic correctness, style consistency, proxy visual realism, and multi-view consistency, using powerful pretrained models to capture what simple pixel metrics cannot.

\paragraph*{Human Perceptual Evaluation (User Studies).}
Given the inherent ambiguities of the task, the ultimate assessment of texture quality often relies on human perception. Many surveyed works include user studies to compare perceptual quality and user preference~\cite{perla2024easitex,richardson2023texture,chen2023text2tex}. Despite the variety of quantitative metrics, user studies remain essential for evaluating perceptual realism and semantic satisfaction, though they are subjective and not perfectly reproducible. Common strategies to reduce bias include diversifying participants and evaluation tasks, randomizing presentation order, and carefully controlling comparison protocols. Typical study designs include pairwise A/B preference tests reporting the \emph{win rate}~($\uparrow$), ranking tasks reporting the \emph{average rank}~($\downarrow$), and absolute rating studies yielding a \emph{mean opinion score (MOS)}~($\uparrow$)~\cite{ITU-R-BT.500-14}.

User studies are typically reported across multiple aspects: \emph{overall quality}; \emph{structural guidance alignment} (respecting mesh semantics during texturing)~\cite{xie2024styletex}; \emph{stylistic guidance alignment}; \emph{level of detail} and \emph{presence of artifacts}~\cite{cao2023texfusion}; \emph{seam visibility}; \emph{diversity}; and \emph{3D consistency} across views~\cite{liu2024syncmvd}. In practice, user studies are most informative when interpreted alongside quantitative metrics.

In a field where no single metric is sufficient, a multifaceted evaluation strategy that combines distribution-level statistics, per-instance accuracy, semantic alignment, user studies, runtime analysis, and robustness tests across different inputs offers the most holistic view of a method’s performance. Each class of metrics captures a distinct aspect of the textured output, and together they reveal the complementary strengths and trade-offs of neural 3D texturing approaches in the literature.

\section{Applications}
\label{sec:applications}

Neural 3D mesh texturing supports a broad range of task settings and downstream applications. Research has progressively addressed complementary problem settings, including synthesizing textures from text~\cite{richardson2023texture,chen2023text2tex} or images~\cite{perla2024easitex}, estimating physically based (PBR) materials~\cite{chen2022tango,zhang2024dreammat}, aligning category-level UVs~\cite{chen2022auvnet}, completing missing appearance on partial scans~\cite{chibane2020implicit}, localized texturing~\cite{decatur2024paintbrush}, and scene-scale texturing~\cite{wang2024roomtex,huang2025roompainter,yang2024instancetex}. Beyond benchmarking, these advances map naturally to practical use cases across content creation, telepresence, visualization, simulation, gaming, and fabrication, where the final output is often a UV-parameterized mesh---and increasingly a full PBR material stack---whose appearance can be edited, relit, and rendered efficiently (Fig.~\ref{fig:appl}):

\begin{figure}[t]
  \centering
  \includegraphics[width=0.99\linewidth]{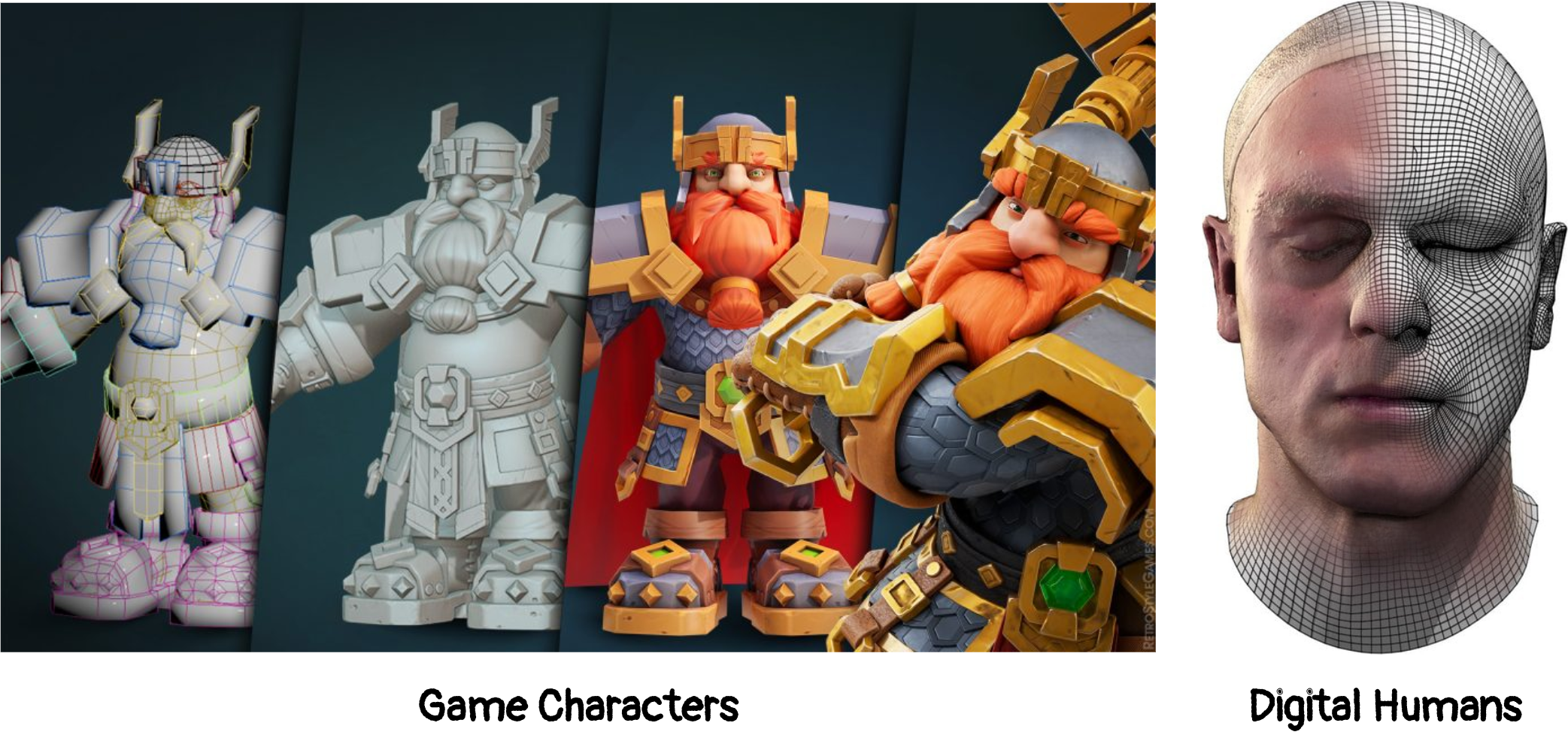}
  \caption{
  Applications of 3D mesh texturing in games and VFX. Textures add material detail and realism to production assets such as game characters (left) and digital humans (right). Figure adapted from~\cite{demianenko2024howtomake3dmodels, wolfe2017scanlineziva}.
  }
  \label{fig:appl}
\end{figure}

\begin{itemize}
  \item \textbf{Asset creation for games, XR, and robotics.} 
  Text- and image-guided pipelines generate high-quality textures on fixed meshes from high-level prompts or exemplars, accelerating look development, style exploration, and re-skinning; representative systems include prompt-driven and diffusion-guided UV synthesis~\cite{richardson2023texture,chen2023text2tex,cao2023texfusion,liu2024syncmvd}. Outputs integrate with real-time engines via PBR material maps (albedo/roughness/normal, \textit{etc.}), enabling fast relighting and consistent deployment across platforms~\cite{chen2022tango,chen2023fantasia3d}. In practice, teams can use these models to produce on-brand variants (\textit{e.g.}, seasonal skins) while preserving decals and layout fidelity inherited from the mesh UVs~\cite{richardson2023texture,chen2023text2tex}. More broadly, diversified textures on scene meshes are often used in simulation to improve sim-to-real transfer for perception and control in robotics~\cite{tobin2017dr}.
 
  \item \textbf{Digital humans and telepresence.}
  Neural textures for deformable meshes offer temporally stable, high-fidelity avatar rendering, reenactment, and appearance editing, making them relevant to telepresence and virtual production~\cite{thies2019deferred}. Related pipelines transfer garment patterns and prints to clothed-body meshes while respecting canonical UVs or part correspondences, supporting virtual try-on and fashion prototyping~\cite{nam2025parte,zhang2024fabricdiffusion}. When PBR is required, texturing methods that estimate SVBRDF channels enable realistic relighting and integration into real or virtual environments~\cite{chen2022tango,zhang2024dreammat}.
  
  \item \textbf{E-commerce and product visualization.}
  For product visualization, texturing techniques enable fabric, leather, or finish swaps on a single mesh, allowing customized previews and photoreal variants. Diffusion- and image-conditioned approaches can reduce manual authoring time while preserving brand constraints and UV semantics~\cite{perla2024easitex,yeh2024texturedreamer}.
  
  
  %
  
  
  \item \textbf{3D stylization and look development.}
  Optimization and feed-forward methods that modify appearance directly on 3D surfaces or meshes enable consistent, multi-view stylization (\textit{e.g.}, painterly, toon, or brand-specific styles) with artist-in-the-loop control; coupling semantic guidance (text/image) with structural cues (normals/depth) improves fidelity and geometry awareness for interactive authoring~\cite{michel2022text2mesh,hollein2022stylemesh,perla2024easitex}.
\end{itemize}

\section{Limitations}
\label{sec:limitations}

While 3D mesh texturing has advanced rapidly with the advent of diffusion-driven generative pipelines, several important challenges remain:

\begin{itemize}
    \item \textbf{Incomplete texturing from multi-view projection.} Methods that texture a mesh by iteratively projecting and inpainting from multiple rendered views---e.g., TEXTure, Text2Tex, and TexFusion---are effective and convenient because they leverage off-the-shelf, pre-trained text-to-image diffusion models and require little or no task-specific training. However, they inherently cover only camera-visible regions in each step; self-occluded or rarely visible areas may remain underspecified and require post-hoc fixes. Even with heuristics for next-best-view scheduling and partial texture masks, view-to-view inconsistency and residual holes are common failure modes \cite{richardson2023texture,chen2023text2tex,cao2023texfusion,liu2024syncmvd}. Recent formulations that synchronize multi-view denoising improve global consistency, but do not fully eliminate occlusion-driven gaps on challenging geometry \cite{liu2024syncmvd}.
    
    \item \textbf{Small and imperfect open-source datasets for \emph{textured} assets.} Large-scale 3D repositories (e.g., Objaverse/Objaverse-XL) provide breadth and diversity, yet textures are heterogeneous in quality, licensing, and parameterization, and rarely include physically based (PBR) channels at scale \cite{deitke2023objaverse,deitke2023objaversexl}. Dedicated material datasets exist---from measured SVBRDFs to modern CC0 PBR collections---but they typically provide \emph{2D} material assets (planar patches) rather than curated, high-quality, per-\emph{mesh} UV textures suitable for training texturing models end-to-end \cite{ma2023opensvbrdf,vecchio2024matsynth}. As a result, supervision remains limited for learning high-fidelity, mesh-aligned appearance.
    
    \item \textbf{Generalization and factorization of appearance.} Models trained on small or weakly curated datasets can bake view-dependent effects (specularities, shadows) into albedo and struggle to predict physically meaningful material maps (albedo/roughness/metalness) that generalize across objects and lighting. Progress in neural inverse rendering and reflectance decomposition underscores both the promise and the difficulty of robust, disentangled estimation under unknown illumination and complex geometry \cite{zhang2021nerfactor,boss2021nerd,ma2023opensvbrdf}. Foundational, broadly generalizable material estimators for in-the-wild meshes remain an open problem.
    
    \item \textbf{Computational cost and memory footprint.} High-resolution UVs (multi-UDIM, $8$--$16$\,K) and multi-view diffusion sampling are expensive in time and GPU memory. Although distillation and consistency-model techniques offer a path to reducing sampling steps, pipelines that couple high-resolution UV baking with multi-view regularization are still resource-intensive, especially in academic settings \cite{luo2023lcm,cao2023texfusion,chen2023text2tex}.
\end{itemize}

\section{Conclusion and Future Work}
\label{sec:future-work}

Neural 3D mesh texturing has rapidly developed into a vibrant research area, with growing impact on 3D asset creation for VFX, e-commerce, advertising, gaming, and related industries. Recent advances in diffusion models, vision--language priors, and differentiable rendering have significantly expanded the capabilities of automated texture generation, enabling workflows that are more scalable, expressive, and less reliant on manual intervention.

In this survey, we presented a comprehensive overview of this field, categorizing methods into foundational neural approaches, optimization-based methods, and accelerated diffusion-based pipelines, while analyzing their design choices in terms of supervision, guidance, and architectural patterns. Alongside a review of current datasets, evaluation strategies, and practical applications, we identified key limitations related to occlusion-aware synthesis, dataset quality, the generalization and factorization of physical appearance, and computational cost.

Looking ahead, future work in neural mesh texturing will likely expand along multiple directions, including greater scale, dynamic and deformable settings, and continued improvements in quality, controllability, and interactivity, alongside related advances in part-aware 3D understanding~\cite{perla2025asia}. We outline a few concrete problems below:


\begin{itemize}
    \item \textbf{Texturing 3D scenes.} Compared with individual 3D objects, scenes exhibit greater geometric diversity and looser structural relationships among constituent objects. This added complexity makes scene-level texturing more challenging, particularly when enforcing appearance consistency across objects, materials, and spatial context.

    \item \textbf{Geometry-aware view planning and coverage.} Beyond heuristic camera schedules, optimization- or learning-based view planning that explicitly maximizes surface coverage and uncertainty reduction could mitigate self-occlusion while minimizing redundant renderings. Integrating coverage metrics with synchronized multi-view denoising may further reduce inconsistency~\cite{chen2023text2tex,liu2024syncmvd}.

    \item \textbf{Dynamic textures and materials.} With video diffusion models maturing, generating temporally coherent animated textures (e.g., flowing patterns or wear over time) is a natural extension. Ensuring temporal stability across UV seams and under motion, including deforming meshes, poses new challenges in conditioning, regularization, and evaluation~\cite{khachatryan2023text2videozero}.

    \item \textbf{Deformation- and correspondence-aware texturing for dynamic meshes.} Extending static pipelines to articulated or topology-varying meshes requires correspondence-robust mapping across frames or poses and appearance transport that respects stretch, compression, and contact. Combining deformation-aware parameterization with multi-view diffusion priors is a promising direction.

    \item \textbf{Scalability and speed.} Practical adoption benefits from faster sampling (few-step or distilled samplers), tile- and patch-based UV generation with seamless blending, and memory-efficient training and inference for multi-UDIM assets. Recent latent consistency approaches suggest a promising path toward substantial speedups while preserving fidelity~\cite{luo2023lcm}.

    \item \textbf{Faithful material decomposition and relightability.} Joint estimation of mesh-aligned albedo, normal, roughness, metallicity, and environment illumination remains difficult at scale. Hybrid pipelines that couple diffusion priors with physically motivated inverse rendering, and that train against measured or high-quality synthetic SVBRDF corpora, could yield more robust and relightable assets~\cite{zhang2021nerfactor,boss2021nerd,ma2023opensvbrdf}.
\end{itemize}

Last but not least, there is still a pressing need for curated, rights-cleared benchmarks of textured \emph{meshes} with high-resolution UVs and, ideally, PBR channels, paired with standardized metrics for multi-view consistency, seam visibility, and perceptual quality under novel lighting and viewpoints. Recent dataset efforts provide important building blocks, but they do not yet provide mesh-aligned supervision at scale~\cite{deitke2023objaverse,ma2023opensvbrdf,vecchio2024matsynth}.


\bibliographystyle{eg-alpha-doi} 
\bibliography{main}


\end{document}